\documentclass[twocolumn]{svjour3}          
\smartqed  

\usepackage{hyperref}
\usepackage{color}
\usepackage{soul}
\usepackage{graphicx}
\usepackage{caption}
\usepackage{subcaption}
\usepackage{amsmath}
\usepackage{amssymb}
\usepackage{algorithm,algorithmic}
\usepackage{paralist}
\usepackage{multirow}

\usepackage[numbers,sort&compress]{natbib}

\newcommand{\R}{\mathbb R}

\newcommand{\RRS}{\R^2\!\times\! S^1}

\journalname{J Math Imaging Vis}
\begin{document}

\title{A neuro-mathematical model for geometrical optical illusions
}


\author{B. Franceschiello         \and
        A. Sarti  \and G. Citti 
}


\institute{B. Franceschiello \at Centre d'analyse et math\'{e}matiques sociales, CNRS - EHESS
              Paris, France \\
              \email{benedetta.franceschiello@ehess.fr}           
           \and
           A. Sarti \at Centre d'analyse et math\'{e}matiques sociales, CNRS - EHESS
           Paris, France \\
           \email{alessandro.sarti@ehess.fr}
                         \and
                         G. Citti \at
                            Dipartimento di Matematica, Universit\`{a} di Bologna \\
                            Bologna, Italy \\
                            \email{giovanna.citti@unibo.it}
                           }

\date{Received: date / Accepted: date}

\maketitle
\begin{abstract}
Geometrical optical illusions have been object of many studies due to the possibility they offer to understand the behaviour of low-level visual processing. They consist in situations in which the perceived geometrical properties of an object differ from those of the object in the visual stimulus. 
Starting from the geometrical model introduced by Citti and Sarti in \cite{CS1}, we provide a  mathematical model and a computational algorithm which allows to interpret these phenomena and to qualitatively reproduce the perceived misperception.
\keywords{Geometrical optical illusions \and Neuro-mathematical model \and Visual cortex \and Infinitesimal strain theory \and Hering illusion}
\end{abstract}

\section{Introduction}
\label{intro}
\begin{figure*}
\centering
\begin{subfigure}[b]{1.8 in}
\centering
 \includegraphics[width= 0.85\columnwidth]{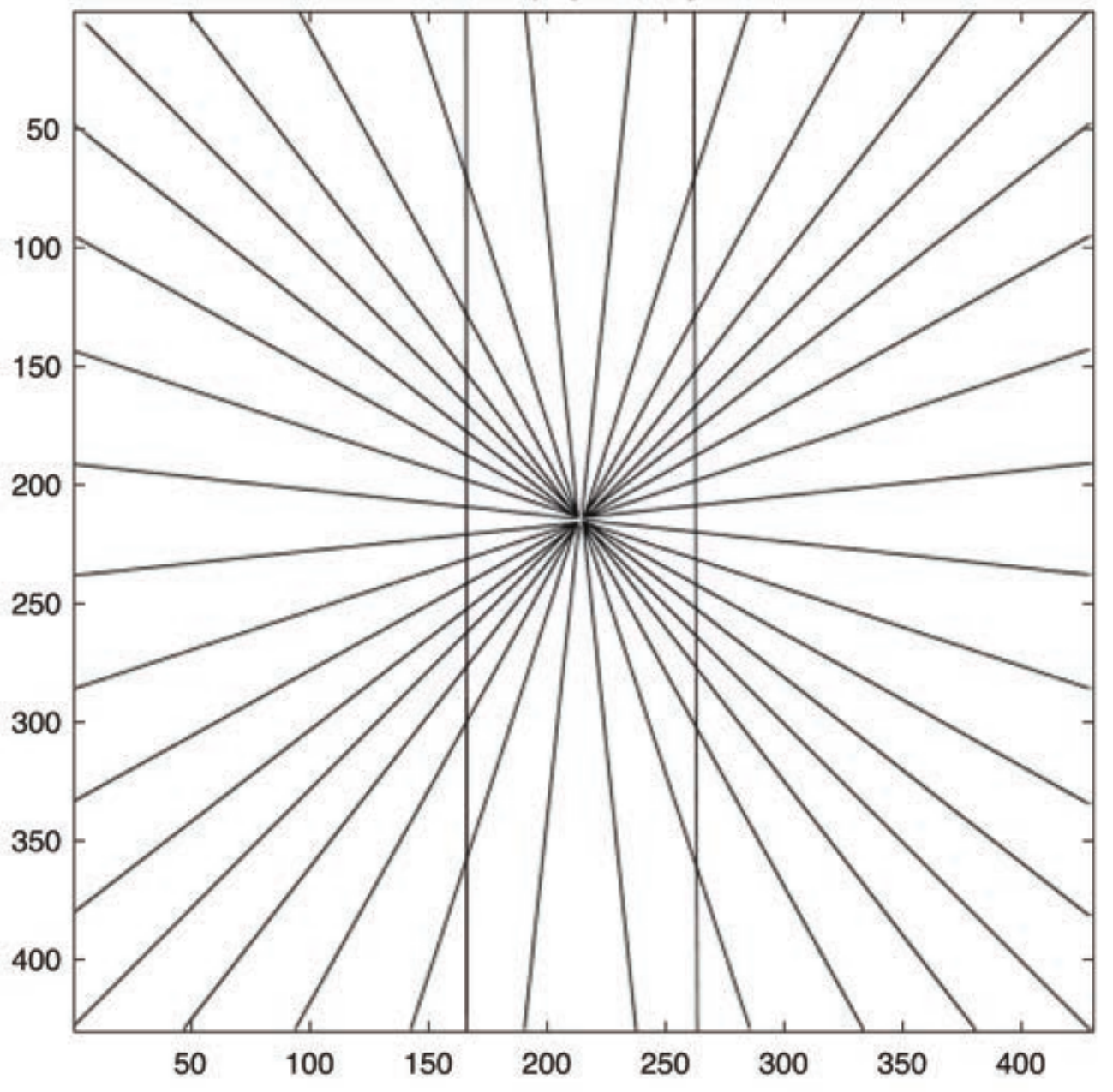}
  \caption{Hering illusion}
  \label{fig:1:4}
 \end{subfigure}
\begin{subfigure}[b]{2.1 in}		
\centering
\includegraphics[width = \columnwidth]{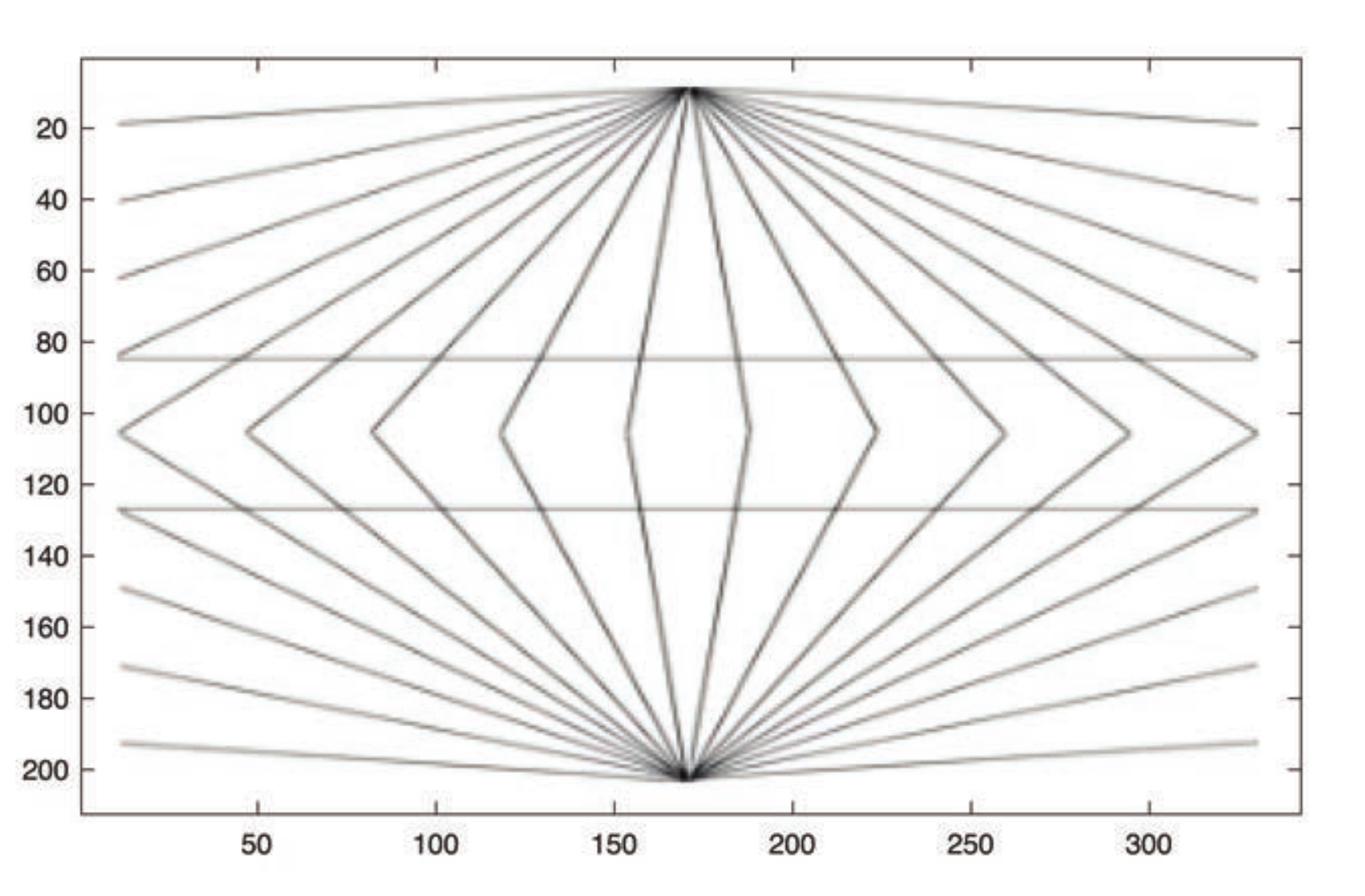}  \caption{Wundt Illusion}
  \label{fig:1:5} 
	\end{subfigure} 
	   	   	 	   	 	\begin{subfigure}[b]{1.8 in}		
	   	   	 	   	 	\centering
	   	   	 	   	 		\includegraphics[width =0.9 \columnwidth]{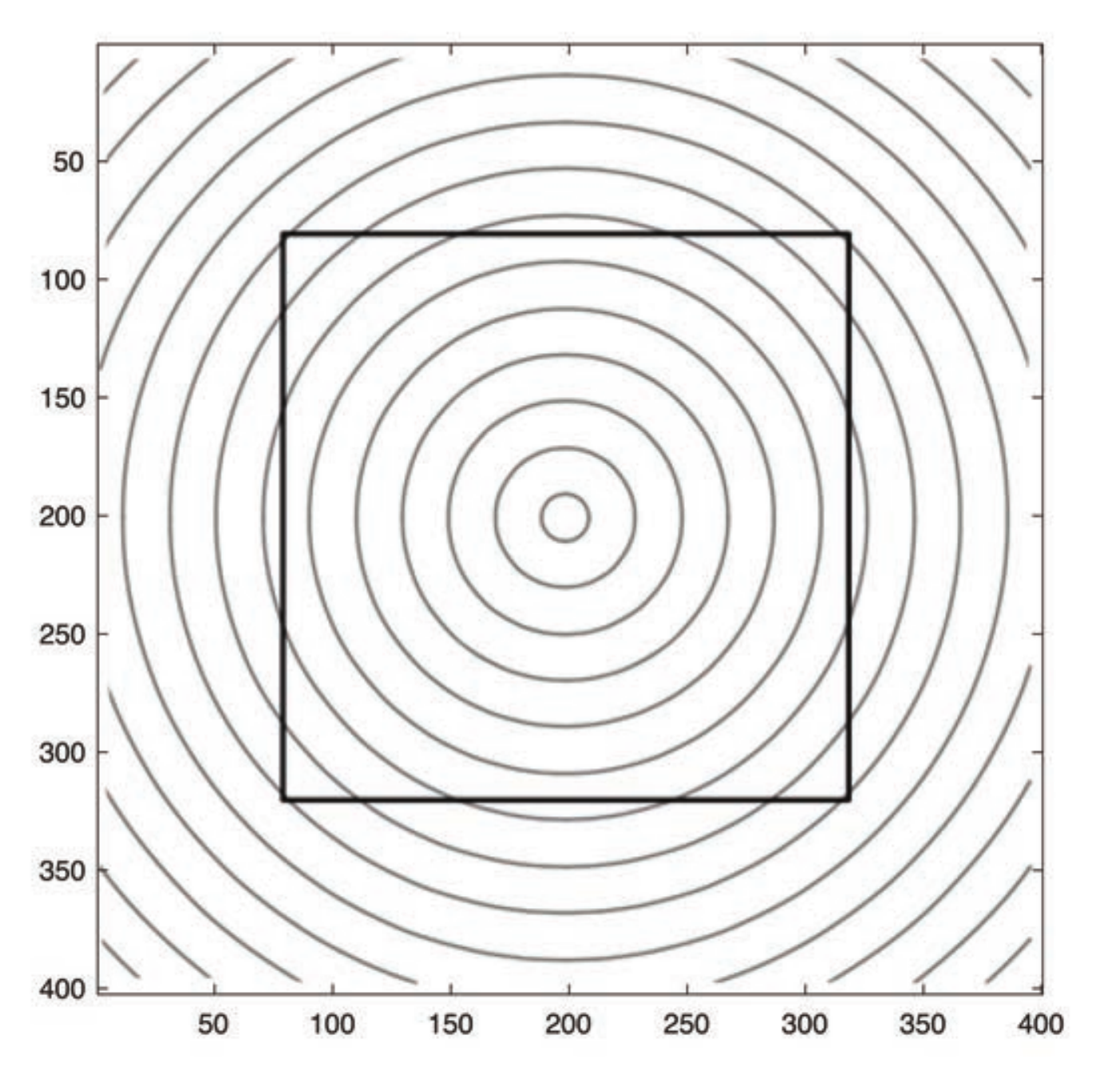}  \caption{Square in Ehrenstein context}
	   	   	 	   	 		\label{fig:1:1} 
	   	   	 	   	 	\end{subfigure}
\begin{subfigure}[b]{2.1 in}		
\centering
\includegraphics[width = \columnwidth]{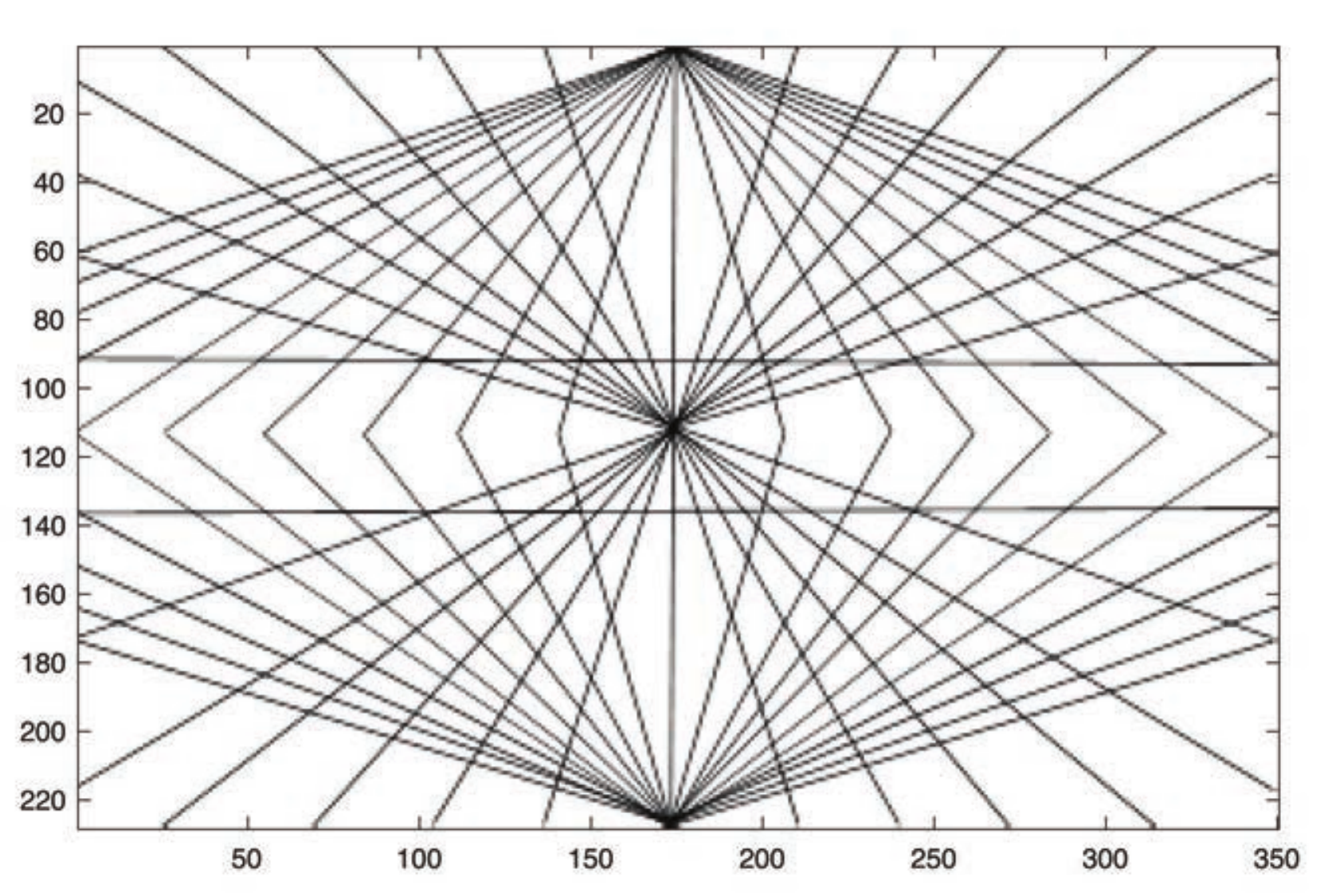}  \caption{Wundt-Hering illusion}
   	\label{fig:1:6} 
\end{subfigure}
   	\begin{subfigure}[b]{1.8 in}
   	\centering
   		\includegraphics[width= 0.5\columnwidth]{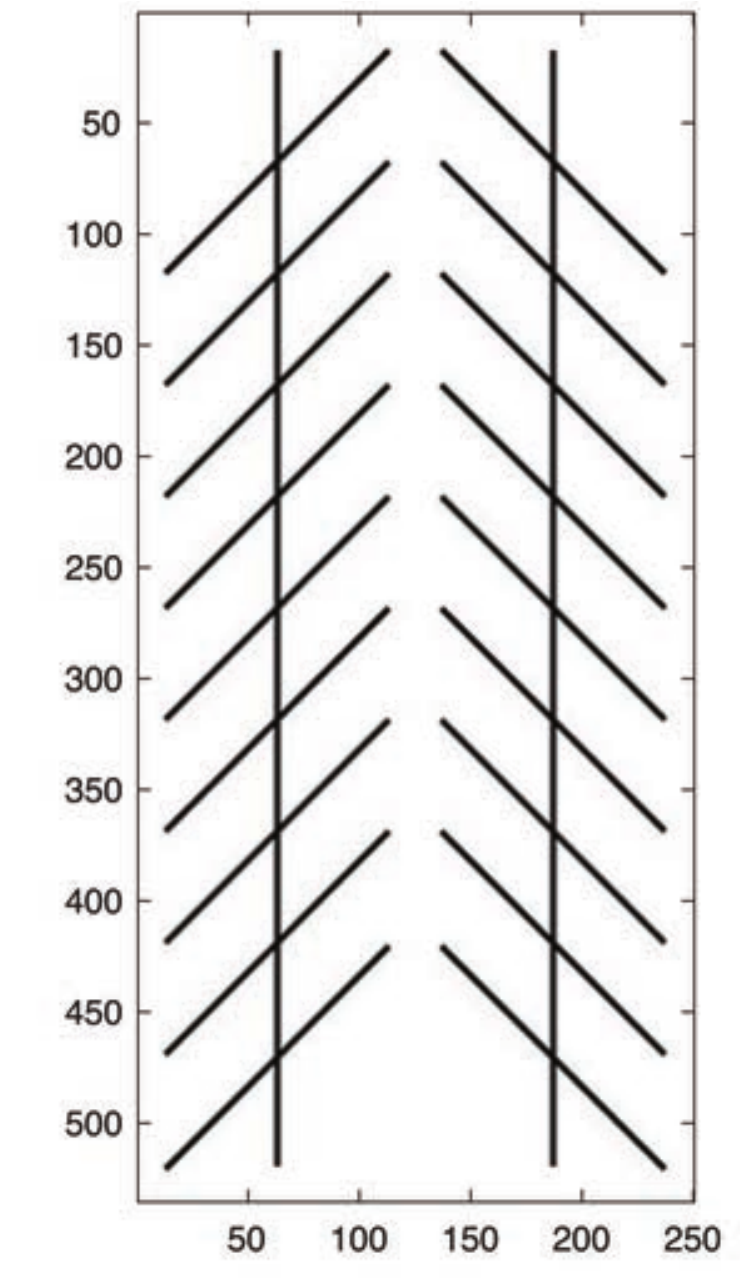} \caption{Zollner illusion}
   		\label{fig:1:2}
   	\end{subfigure}

 	\caption{
 		\subref{fig:1:4} Hering illusion: the two vertical lines are straight and parallel, but since they are presented in front of a radial background the lines appear as if they were bowed outwards. \subref{fig:1:5} Wundt-Illusion: the two horizontal lines are both straight, but they look as if they were bowed inwards. \subref{fig:1:1} Square shape over Ehrenstein context: the context of concentric circles bends the edges of the square toward the center of the image. 
 		 \subref{fig:1:6} Wundt-Hering illusions merged together: the horizontal lines are straight and parallel and the presence of inducers which bow them outwards and inwards at the same time inhibits the bending effect. \subref{fig:1:2} Zollner illusion: a pattern of oblique inducers surrounding parallel lines creates the illusion they are unparallel}
 		 \label{fig:1ills}
 \end{figure*}

Geometrical-optical illusions (GOIs) have been discovered in the XIX century by German psychologists (Oppel 1854 \cite{oppel1855uber}, Hering, 1878, \cite{Her_1}) and have been defined as situations in which there is an awareness of a mismatch of geometrical properties between an item in object space and its associated percept \cite{westheimer2008illusions}. The distinguishing feature of these illusions is that they relate to misjudgements of geometrical properties of contours and they show up equally for dark configurations on a bright background and viceversa. For the interested reader, a historical survey of the discovery of geometrical-optical illusions is included in Appendix I of \cite{westheimer2008illusions}. Our intention here is not to make a classification of these phenomena, which is already widely present in literature (Coren e Girgus, 1978, \cite{coren1978seeing}; Robinson, 1998, \cite{robinson2013psychology}; Wade, 1982, \cite{wade1982art}).
The aim of this paper is to propose a mathematical model for GOIs based on the functional architecture of low level visual cortex (V1/V2). This neuro-mathematical model will allow us to interpret at a neural level the origin of GOIs and to reproduce the arised percept for this class of phenomena. The main idea is to adopt the model of the functional geometry of V1 provided in \cite{CS1} and to consider that the image stimulus will modulate the connectivity. When projected onto the visual space, the modulated connectivity gives rise to a Riemannian metric which is at the origin of the visual space deformation. The displacement vector field at every point of the stimulus is mathematically computed by solving a Poisson problem and the perceived image is finally reproduced. The considered phenomena consist, as shown in figure \ref{fig:1ills}, in straight lines over different backgrounds (radial lines, concentric circles, etc). The interaction betwen target and context either induces an effect of curvature of the straight lines (fig. \ref{fig:1:4}, \ref{fig:1:5}, \ref{fig:1:1}), eliminates the bending effect (fig. \ref{fig:1:6}), or induces an effect of unparallelism (fig. \ref{fig:1:2}).
The paper is organised as follows: in section \ref{sec:1} we review the state of the art concerning the previous mathematical models proposed. In section \ref{sec:2} we will briefly recall the functional achitecture of the visual cortex and the cortical based model introduced by Citti and Sarti in \cite{CS1}. In section \ref{sec:3bis} we will introduce the neuro-mathematical model proposed for GOIs, taking into account the modulation of the functional architecture induced by the stimulus. In \ref{sec:3} the numerical implementation of the mathematical model will be explained and applied to a number of examples. Results are finally discussed.

\section{Geometrical optical illusions}
\label{sec:1}
\subsection{Role of GOIs}
\label{sec:11}
 In psychology the distal stimulus is defined as \textit{the light reflected off a physical object in the external world}; when we look at an image (distal stimulus) we cannot actually experience the image physically with vision, we can only experience it in our mind as proximal stimulus \cite{koffka2013principles}, \cite{gibson1960concept}. 
Geometrical optical illusions arise when the distal stimulus and its percept differ in a perceivable way. 
As explained by Westheimer in \cite{westheimer2008illusions}, we can conveniently divide illusions into those in which spatial deformations are a consequence of the exigencies of the processing in the domain of brightness and the true geometrical-optical illusions, which are misperceptions of geometrical properties of contours in simple figures. Some of the most famous geometric illusions of this last type are shown in figure \ref{fig:1ills}.
The importance of this study lies in the possibility, through the analysis of these phenomena combined with physiological recordings, to help to guide neuroscientific research (Eagleman, \cite{eagleman2001visual}) in understanding the role of lateral inhibition, feedback mechanisms between different layers of the visual process and to lead new experiments and hypothesis on receptive fields of V1 and V2. 
Many studies, which relies on neuro-physiological and imaging data, show the evidence that neurons in at least two visual areas, V1 and V2, carry signals related to illusory contours, and that signals in V2 are more robust than in V1 (\cite{von1984illusory}, \cite{murray2002spatiotemporal}, reviews \cite{eagleman2001visual}, \cite{murray2013illusory}). A more recent study on the tilt illusion, in which the perceived orientation of a grating differs from its physical orientation when surrounded by a tilted context, measured the activated connectivity in and between areas of early visual cortices (\cite{song2013effective}). These findings suggest that for GOIs these areas may be involved as well.
Neurophysiology can help to provide a physical basis to phenomenological experience of GOIs opening to the possibility of mathematically modeling them and to integrate subjective and objective experiences. 

\subsection{Mathematical models proposed in literature}
\label{sec:12}
The pioneering work of Hoffman  \cite{hoffman1971visual} dealt with illusions of angle (i.e. the ones involving the phenomenon of angular expansion, which is the tendence to perceive under certain conditions acute angles as larger and obtuse ones as smaller) modeling the generated perceived curves as orbits of a Lie transformation group acting on the plane. The proposed model allows to classify the perceptual invariance of the considered phenomena in terms of Lie Derivatives, and to predict the slope. Another model mathematically equivalent to the one proposed by Hoffman has been proposed by Smith, \cite{smith1978descriptive}, who stated that the apparent curve of geometrical optical illusions of angle can be modeled by a first-order differential equation depending on a single parameter. By computing this value an apparent curve can be corrected and plotted in a way that make the illusion being not perceived anymore (see for example fig. 8 of \cite{smith1978descriptive}). This permits to introduce a \textit{quantitative} analysis of the perceived distortion.
Ehm and Wackerman in \cite{ehm2012modeling}, started from the assumption that GOIs depend on the context of the image which plays an active role in altering components of the figure. On this basis they provided a variational approach computing the deformed lines as minima of a functional depending on length of the curve and the deflection from orthogonality along the curve. This last request is in accordance to the phenomenological property of regression to right angle. One of the problems pointed out by the authors is that the approach doesn't take into account the underlying neurophysiological mechanisms. 
An entire branch for modeling neural activity, the Bayesian framework, had its basis in Helmholtz’s theory \cite{von2005treatise}: \textit{our percepts are our best guess as to what is in the world, given both sensory data and prior experience.} 
The described idea of unconscious inference is at the basis of the Bayesian statistical decision theory, a principled method for determining optimal performance in a given perceptual task (\cite{geisler2002illusions}). These methods consists in attributing a probability to each possible true state of the environment given the stimulus on the retina and then to establish the way prior experience influences the final guess, the built proximal stimulus (see \cite{knill1996perception} for examples of Bayesian models in perception). An application of this theory to motion illusions has been provided by Weiss et al in \cite{weiss2002motion}, and a review in \cite{geisler2002illusions}.
Ferm\"{u}ller and Malm in \cite{Ferm} attributed the perception of geometric optical illusions to the statistics of visual computations. Noise (uncertainty of measurements) is the reason why systematic errors occur in the estimation of the features (intensity of the image points, of positions of points and orientations of edge elements) and illusions arise as results of errors due to quantization. 
Walker (\cite{walker1973mathematical}) tried to combine neural theory of receptive field excitation together with mathematical tools to provide an equation able to determine the disparity between the apparent line of an illusion and its corresponding actual line, in order to reproduce the perceptual errors that occur in GOIs (the ones involving straight lines).
In our model we aim to combine psycho-physical evidence and neurophysiological findings, in order to provide a neuro-mathematical model able to interpret and simulate GOIs. 

\section{The classical neuromathematical model of V1/V2}\label{sec:2}

\subsection{Neurogeometry of the primary visual cortex}
\label{sec:21}
The visual process is the result of several retinic and cortical mechanisms which act on the visual signal. The retina is the first part of the visual system responsible for the trasmission of the signal, which passes through the Lateral Geniculate Nucleus, where a pre-processing is performed and arrives in the visual cortex, where it is further processed. 
The receptive field (RF) of a cortical neuron is the portion of the retina which the neuron reacts to, and the receptive profile (RP)  $\psi(\xi)$ is the function that models the activation of a cortical neuron when a stimulus is applied to a point $\xi = (\xi_1,\xi_2)$ of the retinal plane. 

\subsubsection{The set of simple cells receptive profiles} \label{211}
\begin{figure}[H]
 \centering
 \includegraphics[width = \columnwidth]{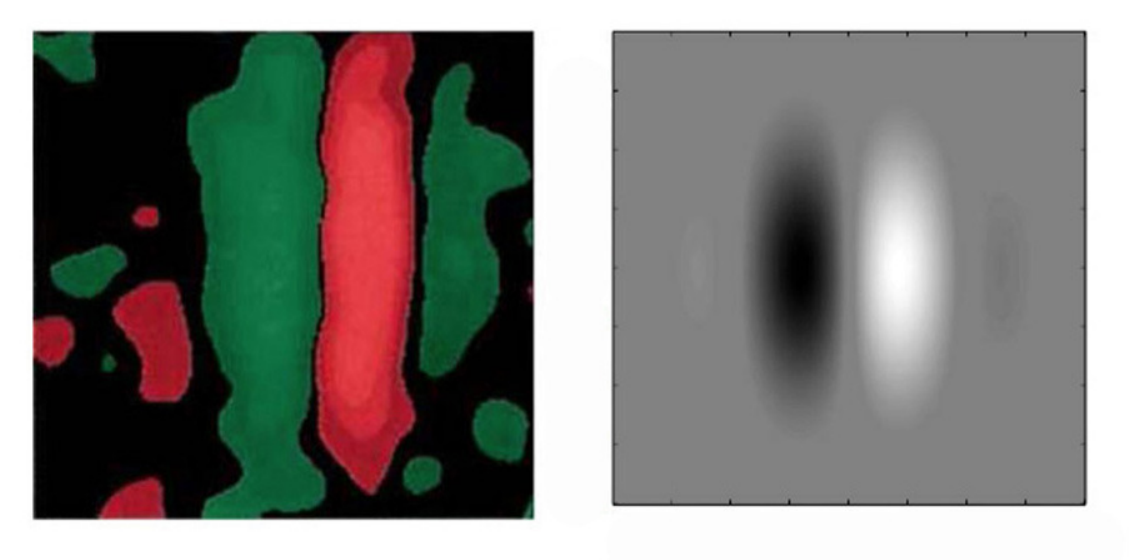}
 \caption{In vivo registered odd receptive field (left, from (De Angelis et al., 1995) \cite{deangelis1995receptive}) and a schematic representation of it as a Gabor filter (right), see \eqref{receptive_profile}}
 \label{fig:2e}
\end{figure}

Simple cells of visual cortices V1 and V2 are sensitive to position and orientation of the contrast gradient of an image. Their properties have been experimentally described by De Angelis in \cite{deangelis1995receptive}, see figure \ref{fig:2e}. From the neurophysiological point of view the orientation selectivity, the spatial and temporal frequency of cells in V2 differs little from the one in V1 (\cite{levitt1994receptive}). Receptive fields in V2 are larger from those in V1 ( \cite{kennedy1985receptive}, \cite{levitt1994receptive}).
Considering a basic geometric model, the set of simple cells RPs can be obtained via translations of vector $x = (x_1, x_2)$ and rotation of angle $\theta$ from a unique mother profile $\psi_0(\xi)$. Daugman \cite{Daug}, Jones and Palmer \cite{jones1987evaluation} showed that Gabor filters were a good approximation for receptive profiles of simple cells in the primary visual cortices V1 and V2. Another approach is to model receptive profiles as Gaussian derivatives, as introduced by Young in \cite{young1987gaussian} and Koenderink in \cite{koenderink1990receptive}, but for our purposes the two approaches are equivalent. A good expression for the mother Gabor filter is:
 \begin{equation}
\psi_0 (\xi) = \psi_0 (\xi_1, \xi_2) = \frac{1}{4 \pi \sigma^2} e^{\frac{-(\xi_1^2 + \frac{\xi_2^2}{4})}{2\sigma^2}}e^{\frac{2i\bar{b}\xi_2}{\sigma}}, 
 \label{receptive_profile}
 \end{equation}
 where $\bar{b}=0.56$ is the ratio between $\sigma$ and the spatial wavelength of the cosine factor.
 
 Translations and rotations can be expressed as:
\begin{equation} \label{group_law}
A_{(x_1,x_2,\theta)}(\xi) = \left(\begin{array}{l}
  x_1 \\ x_2 
\end{array}\right) + \left(\begin{array}{ll}
 \cos\theta &-\sin\theta \\ \sin\theta &\cos\theta 
\end{array}\right) \left(\begin{array}{l}
 \xi_1 \\ \xi_2
\end{array} \right).                 
\end{equation}
Hence a general RP can be expressed as:
\begin{equation*} \psi_{(x_1,x_2,\theta)} (\xi_1, \xi_2) = \psi_0(A^{-1}_{(x_1,x_2,\theta)}(\xi_1, \xi_2)).
  \end{equation*}

\begin{figure}
\centering
    		\includegraphics[width=0.9 \columnwidth]{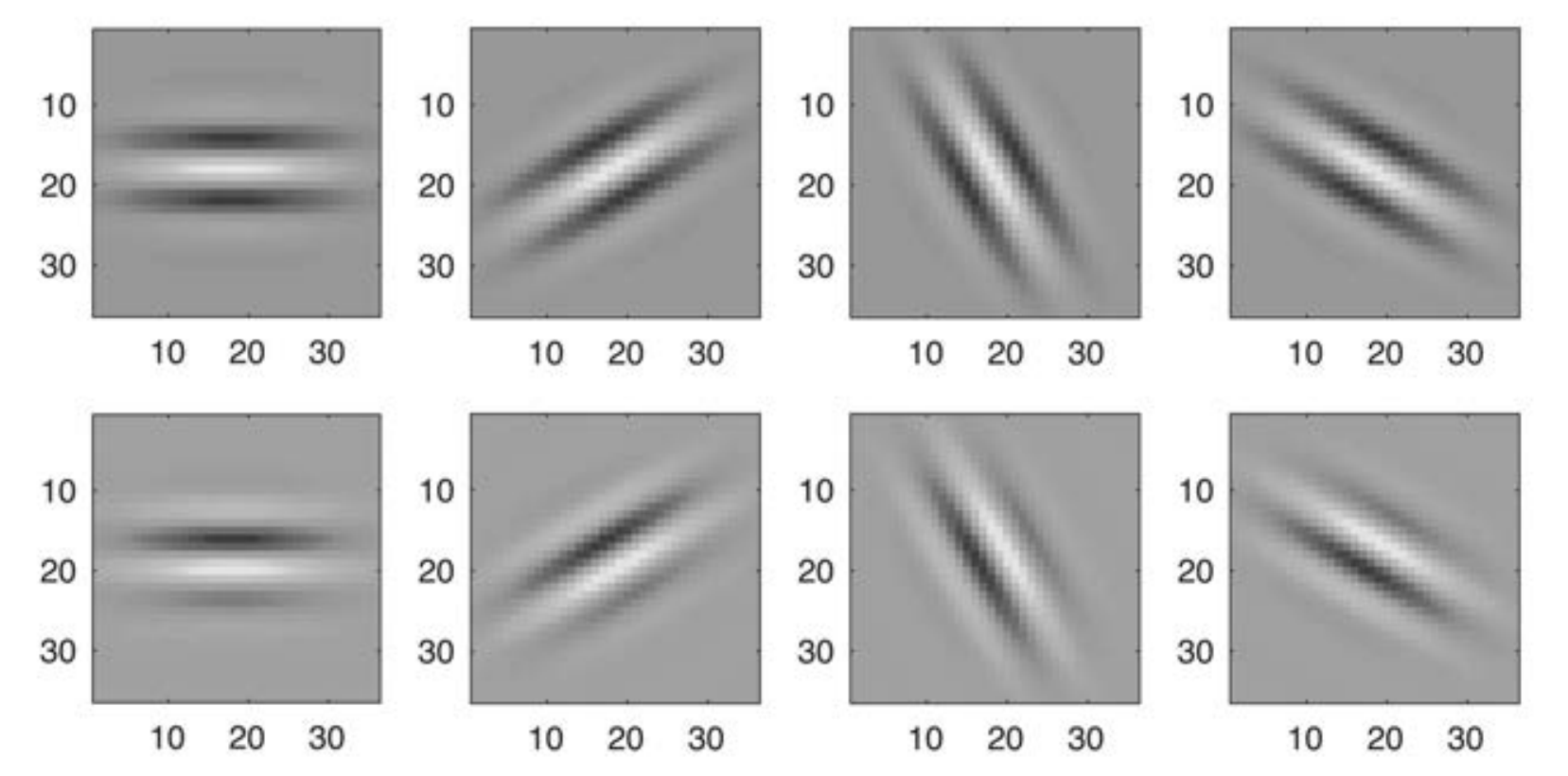} 
    		
  	\caption{ In each image: (top) even part of Gabor filters (real part), (bottom) odd one (imaginary part). Corresponding orientation from left to right: $\theta=0$, $\theta= \pi/6$, $\theta = 2\pi/3$, $\theta = 5\pi/6$, with $\sigma = 4.48$ pixels}
  	\label{fig43332}
\end{figure}

A set of RPs generated with equation \ref{group_law} is shown in figure \ref{fig43332}.

\subsubsection{Output of receptive profiles} 
\label{212}                              

The retinal plane is identified with the $\mathbb{R}^2$-plane, whose local coordinates will be denoted with $x=(x_1,x_2)$. When a visual stimulus $I$ of intensity $I(x_1,x_2) : M \subset \mathbb{R}^2 \rightarrow \mathbb{R}^+$ activates the retinal layer of photoreceptors, the neurons whose RFs intersect $M$ spike and their spike frequencies $O(x_1,x_2,\theta)$ can be modeled (taking into account just linear contributions) as the integral of the signal $I(x_1,x_2)$ with the set of Gabor filters. The expression for this output is: 
\begin{equation}
O(x_1,x_2,\theta) = \int_M \, I(\xi_1, \xi_2)\, \psi_{(x_1,x_2,\theta)}\, (\xi_1, \xi_2) \, d\xi_1 d\xi_2 .
\label{output_simp_cel}
\end{equation}
In the right hand side of the equation the integral of the signal with the real and imaginary part of the Gabor filter is expressed. The two families of cells have different shapes, hence they detect different features. In particular odd cells will be responsible for boundary detection. 

\subsubsection{Hypercolumnar structure}
\label{sec:221}
The term \textit{functional architecture} refers to the organisation of cells in the primary visual cortex in structures. 
The hypercolumnar structure, discovered by the neuro-physiologists Hubel and Wiesel in the 60s (\cite{hubel1977ferrier}), organizes the cells of V1/V2 in columns (called hypercolums) covering a small part of the visual field $M \subset \mathbb{R}^2$ and corresponding to parameters such as orientation, scale, direction of movement, color, for a fixed retinal position $(x_1,x_2)$. 
\begin{figure}[H] 
\centering
\includegraphics[width=0.9\columnwidth]{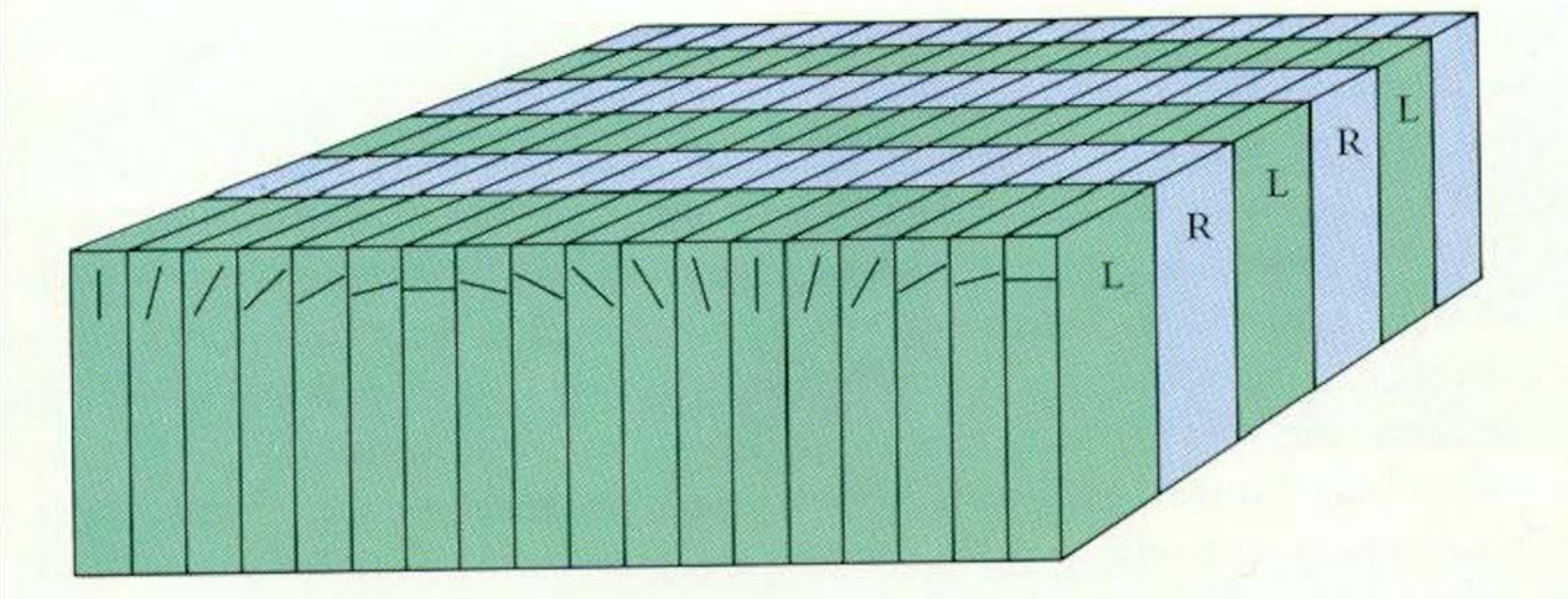}   \includegraphics[width=0.9\columnwidth]{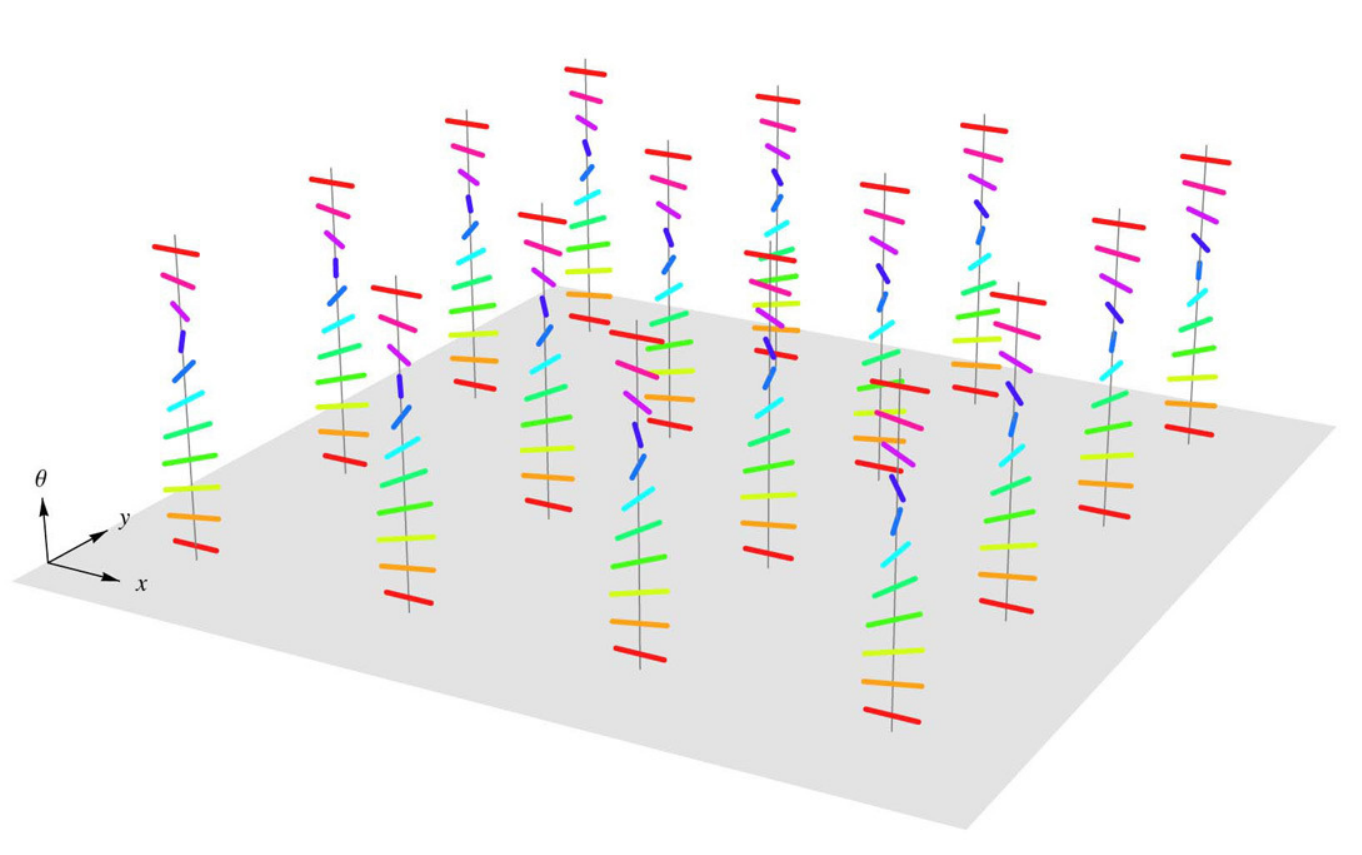}
\caption{ Top: representation of hypercolumnar structure, for the orientation parameter, where L and R represent the ocular dominance columns (Petitot \cite{petitot2008neurogeometrie}). Bottom: for each position of the retina $(x_1,x_2)$ we have the set of all possible orientations}\label{hyper}
\end{figure}
In our framework over each retinal point we will consider a whole hypercolumn of cells, each one sensitive to a specific instance of orientation. Hence for each position $(x_1,x_2)$ of the retina $M \subset \mathbb{R}^2$ we associate a whole set of filters 
\begin{equation}
RP_{(x_1,x_2)} = \{ \psi_{(x_1,x_2,\theta)}:  \,\, \theta \in \mathit{S}^1 \} . 
\end{equation}
 This expression associates to each point of the proximal stimulus in $\mathbb{R}^2$ all possible feature orientations into the space of features $ \mathit{S}^1$, and defines a fiber over each point
 \begin{equation*} \{\theta \in \mathit{S}^1 \}. \end{equation*}
In this way the hypercolumnar structure is described in terms of differential geometry, but we need to explain how the orientation selectivity is performed by the cortical areas in the space of feature $\mathit{S}^1$ (\cite{CS1}).

\subsubsection{Cortical connectivity}
\label{sec:223}
Physiologically the orientation selectivity is the action of short range connections between simple cells belonging to the same hypercolumn to select the most probable response from the energy of receptive profiles.
Horizontal connections are long ranged and connect cells of approximately \textit{the same orientation}.
Since the connectivity between cells is defined on the tangent bundle, we define now the generator of this space.  The change of variable defined through $A$ in \eqref{group_law} acts on the basis for the tangent bundle $(\frac{\partial}{\partial x_1}, \frac{\partial}{\partial x_2})$ giving as frame in polar coordinates:
\begin{equation*}
X_1 = \cos \theta \frac{\partial}{\partial x_1} + \sin \theta \frac{\partial}{\partial x_2}, \,\, X_3= -\sin \theta \frac{\partial}{\partial x_1} + \cos\theta \frac{\partial}{\partial x_2}.
\end{equation*}
As presented in \cite{CS1}, the whole space of features $(x_1,x_2,\theta)$ is described in terms of a 3-dimensional fiber bundle, whose generators are $X_1$, $X_3$ for the base and 
\begin{equation*}
X_2 = \frac{\partial}{\partial \theta},
\end{equation*}
for the fiber. These vector fields generate the tangent bundle of $\RRS$. 

Since horizontal connectivity is very anysotropic, the three generators are weighted by a strongly anysotropic metric.  We introduce now the sub-Riemannian metric with whom Citti and Sarti in \cite{CS1} proposed to endow the $\RRS$ group to model the long range connectivity of the primary visual cortex V1. Starting from the vector fields $X_1$, $X_2$ and $X_3$ we define a metric $g_{ij}$ for which the inverse (responsible for the connectivity in the cortex) is:

\begin{equation} g^{ij}(x_1,x_2,\theta)= \left(\begin{array}{ccc}
      \cos^2\theta  & \sin\theta \cos\theta & 0\\
      \sin\theta\cos\theta  &  \sin^2\theta & 0 \\
      0  &  0 & 1 \\
     \end{array}\right),
\end{equation}
with $i,j=1,2,3$. Cortical curves in V1 will be a linear combination of vector fields $X_1$ and $X_2$, the generators of the 2-dimensional horizontal space, while they will have a vanishing component along $X_3 = [X_1, X_2]$.
The functional architectures built in $\RRS$ correspond to the neural connectivity measured by Angelucci et al. in \cite{angelucci2002circuits} and Bosking et al. in \cite{bosking1997orientation}. For a qualitatively and quantitative comparison between the kernels and the connectivity patterns see Favali et al. in \cite{favali2015local}. In our work a local formulation of the kernel presented in \cite{favali2015local} will be used. 
Furthermore, it has been shown by Sanguinetti et al. in \cite{sanguinetti2010model} that the geometry of fuctional architecture formally introduced in \ref{sec:2} is naturally encoded in the statistics of natural images. Hence these geometrical structures are compatible with Bayesian learning methods.
\section{The neuro-mathematical model for GOIs}
\label{sec:3bis}
\subsection{Output of Simple Cells and connectivity metric} \label{sec:233}
We consider simple cells at fixed value of $\sigma$ depending on position and orientation. 
For fixed value of $(x_1,x_2)$, we restrict the connectivity tensor 
$(g^{ij})_{i,j=1,2,3}$ to the 
$\mathbb{R}^2$ plane, subset of the tangent plane to $\RRS$ at the point $(x_1,x_2,\theta)$, 
and obtain the tensor 
\begin{equation*}
\left(\begin{array}{cc}
      \cos^2\theta  & \sin\theta \cos\theta \\
      \sin\theta\cos\theta  &  \sin^2\theta  \\
     \end{array}\right).
\end{equation*}
For every value of $\theta$ this tensor has only one non zero eigenvalue. The corresponding eigenvector has direction $\theta$. 
We will assign to the norm of the output the usual meaning of energy 
\begin{equation*}
E(x_1,x_2,\theta) = \Arrowvert O(x_1,x_2,\theta)\Arrowvert,
\end{equation*}
where the output is defined in \eqref{output_simp_cel} 
and is evaluated at the fixed value of $\sigma$. We will discuss in section \ref{sec:3} the choice of $\sigma$ for our experiments.
Each point of the hypercolumn is weighted by the energy of simple cells normalized over the whole set of hypercolumn responses:
\begin{equation}
\frac{E(x_1,x_2,\theta)}{\int_{0}^{\pi} E(x_1,x_2,\theta)  d\theta}.
\label{compo_sing}
\end{equation}
The normalization of the output expresses the probability that a specific cell sensitive to $\theta$ within the hypercolumn over $(x_1,x_2)$ is selected. The mechanism of intracortical selection attributing a probability to each possible orientation (state) given the initial stimulus is connected to the long-range activity: simple cells belonging to different hypercolumns in a neighbourhood of a point $(x_1,x_2)$ sensitive to the same orientation will have a high probability.

\begin{figure}
\centering 
\begin{subfigure}[b]{1.8 in}

 \includegraphics[width= \columnwidth]{Fig1a-eps-converted-to.pdf}
  \caption{Hering illusion, distal stimulus}
     		\label{fig:8:1}
  \end{subfigure}
   	\begin{subfigure}[b]{1.8 in}
   		\includegraphics[width= \columnwidth]{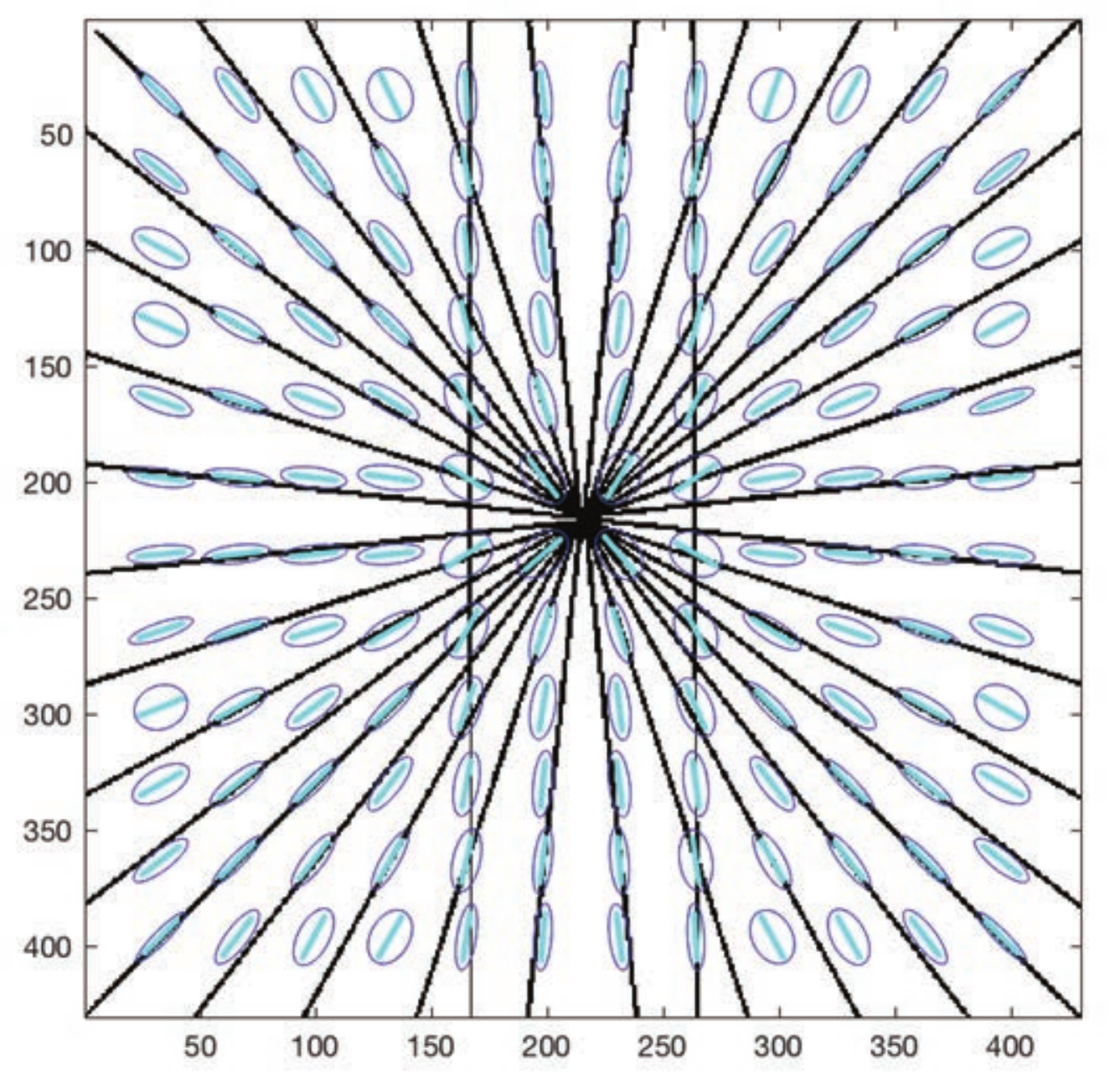} \caption{Tensor representation}
   		\label{fig:8:2}
   	\end{subfigure}
   	   	\begin{subfigure}[b]{1.8 in}
   	   		\includegraphics[width= \columnwidth]{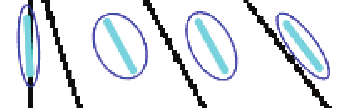} \caption{Detail}
   	   		\label{fig:8:3}
   	   	\end{subfigure}
 	\caption{
 		\subref{fig:8:1} Proximal stimulus (Hering illusion). \subref{fig:8:2} Representation of $\textbf{p}^{-1}$ (blue). Principal and second eigenvectors correspond to first and second semi-axes of the ellipses. Lengths of the semi-axes is given by the magnitude of the corresponding eigenvalues. Principal eigenvectors of ellipses are oriented along the maximum activity registred at $\bar{\theta}$ over each point $(x_1,x_2)$, marked in cyan vector. \subref{fig:8:3} Here we show a detail of the tensor field representation: we notice that along parts of the stimulus strongly oriented, ellipses are elongated. As far as we move further from the level-lines, ellipses lost their elongated form and become rounded, since the stimulus does not respond to a preferred orientation anymore}
 		\label{her_fig}
 \end{figure}

The connectivity tensor restricted to the $\mathbb{R}^2$ plane and modulated by the output of simple cells will become:
\begin{equation}\label{ptensor}
\frac{E(x_1,x_2,\theta)}{\int_{0}^{\pi}  E(x_1,x_2,\theta)  d\theta}
 \left(\begin{array}{cc}
      \cos^2\theta  & \sin\theta \cos\theta \\
      \sin\theta\cos\theta  &  \sin^2\theta  \\
     \end{array}\right). 
\end{equation}
This last expression corresponds to a connectivity polarized by the normalized energy of simple cells shown in \eqref{compo_sing} at points $(x,y,\theta)$. 
The overall cometric (inverse of the metric tensor) arising from the action within the hypercolumn over each retinal point $(x_1,x_2)$ is obtained summing up along $\theta$ the previous modulated metric in \eqref{ptensor} and will have the following expression: 
\begin{equation}
\textbf{p}^{-1}(x_1,x_2)=  \gamma^{-1}\, \frac{\int_{0}^{\pi} E(x_1,x_2,\theta)  \left(\begin{array}{cc}
      \cos^2\theta  & \sin\theta \cos\theta \\
      \sin\theta\cos\theta  &  \sin^2\theta  \\
     \end{array}\right) d\theta}{\int_{0}^{\pi}  E(x_1,x_2,\theta)  d\theta},
     \label{ptensorvero}
\end{equation}

where $ \gamma^{-1}$ is a constant factor. This tensor will have principal eigenvector in the direction $\bar{\theta}$, the orientation corresponding to the maximum energy within the hypercolumn. A visualization of $\textbf{p}^{-1}$ is given in figure \ref{her_fig}. Hence this process describes the selection at every point $(x_1,x_2)$ of the most likely direction of propagation of the connectivity, expressed by the values attained by the energy. 

 \begin{figure*}
  \centering 
 \begin{subfigure}[b]{2.1 in}
  \centering 
  \includegraphics[width= \columnwidth]{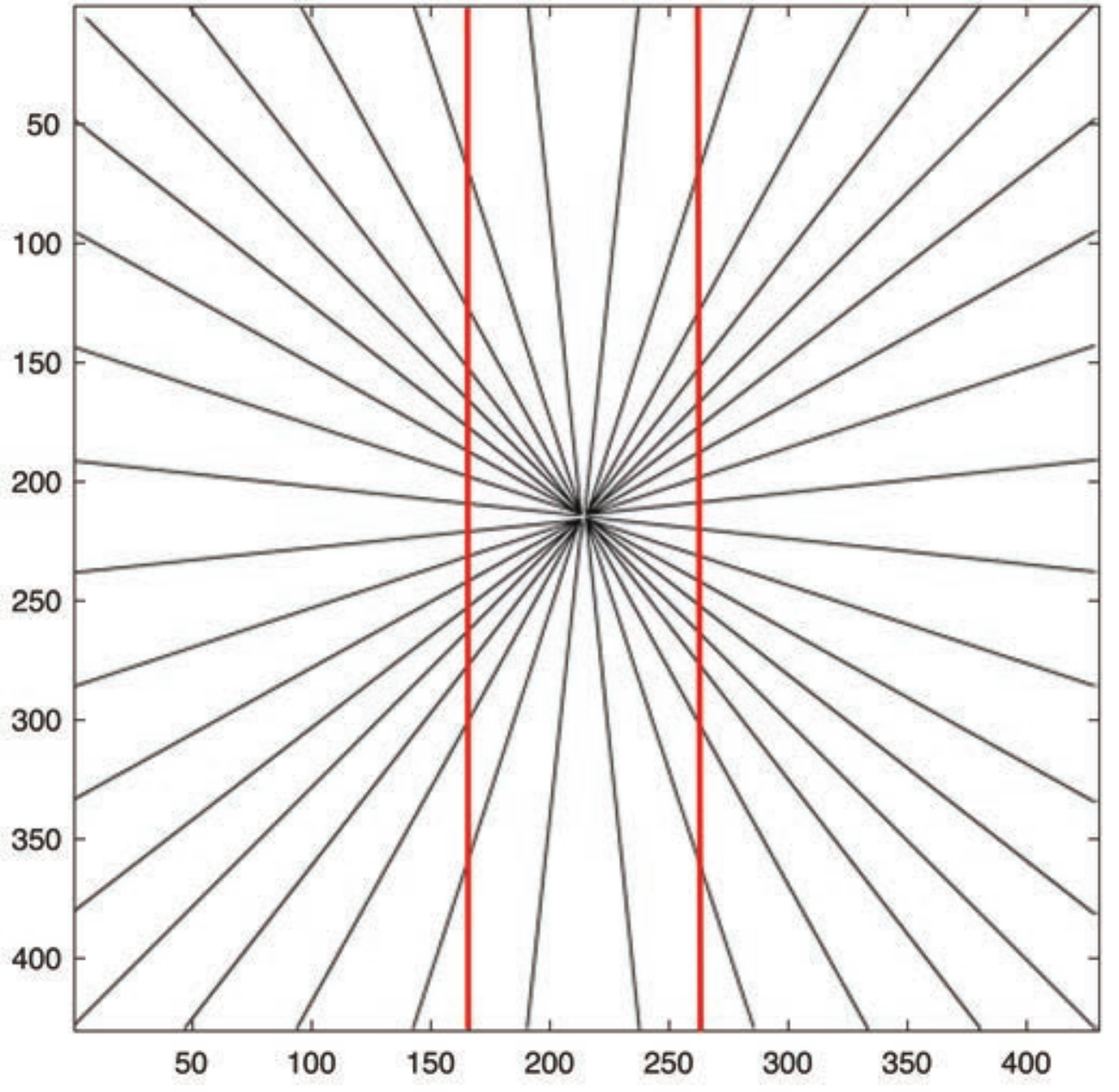}
   \caption{Hering with red lines superimposed}
      	\label{fig:10:1}
   \end{subfigure}
    	\begin{subfigure}[b]{2.1 in}
    		\includegraphics[width= \columnwidth]{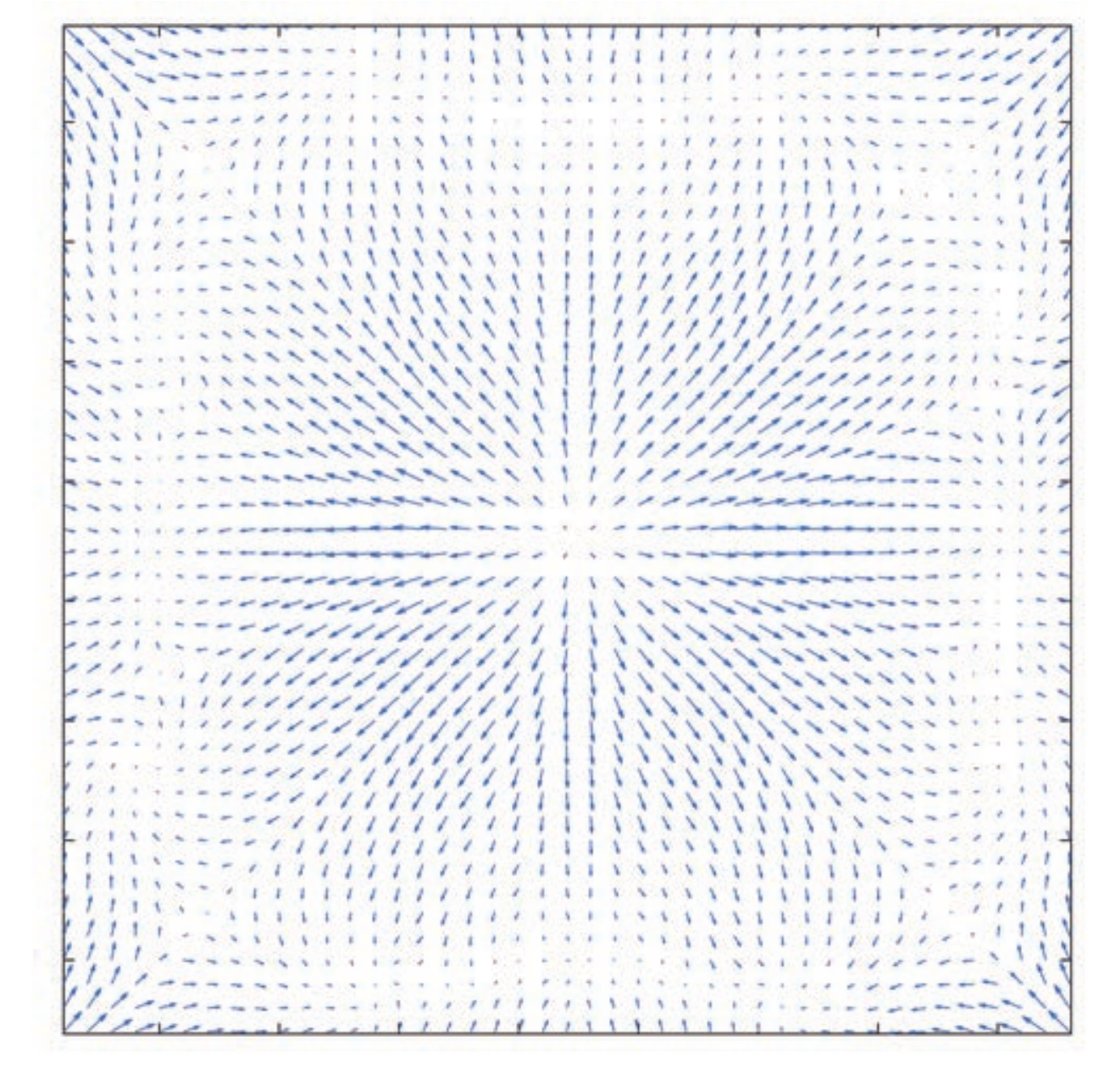} \caption{Displacement vector fields}
    		\label{fig:10:2}
    	\end{subfigure}
\begin{subfigure}[b]{2.1 in}
    	    		\includegraphics[width= \columnwidth]{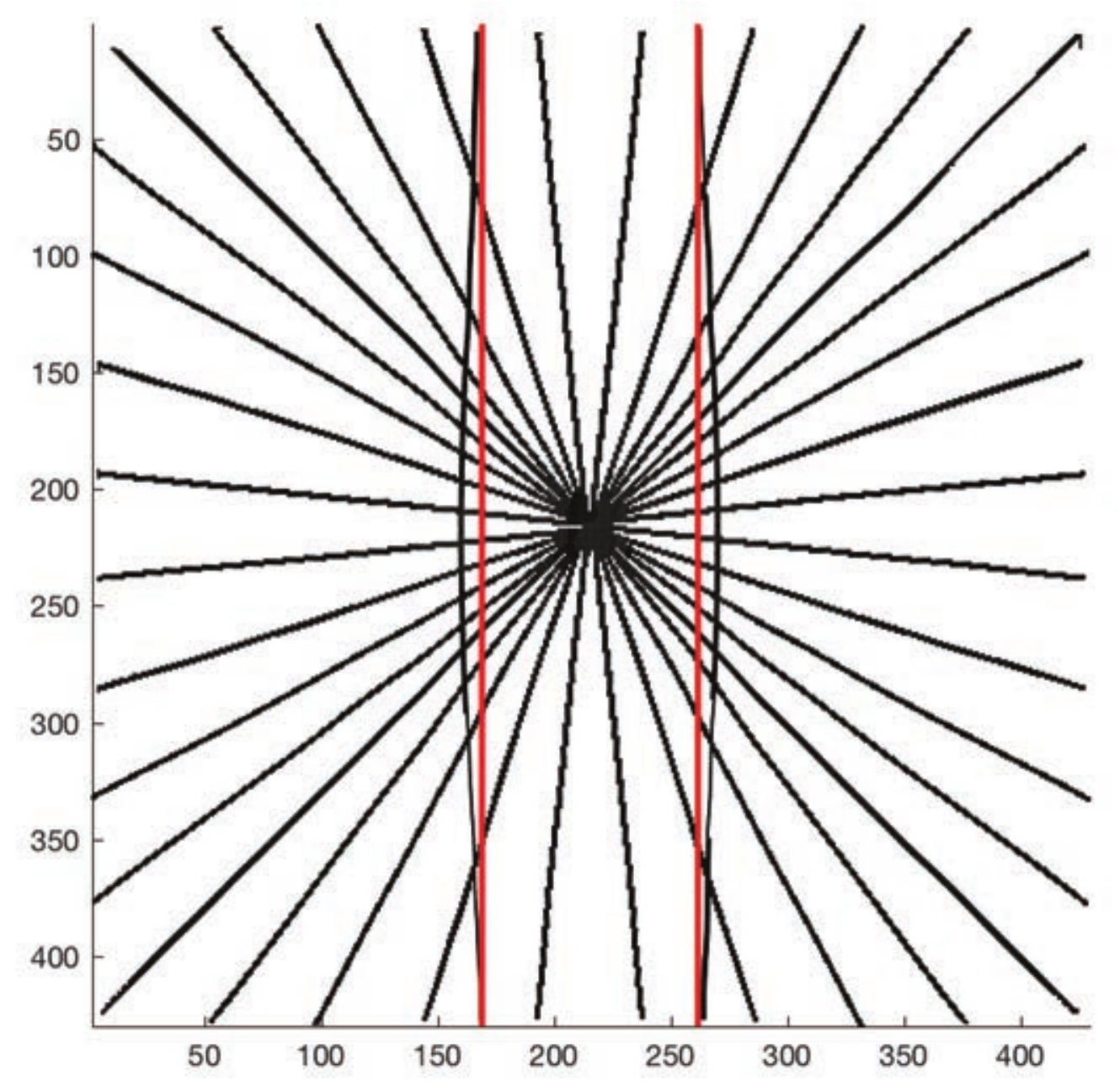} \caption{Proximal stimulus}
    	    	\label{fig:10:3}
\end{subfigure}
    	    	\caption{\subref{fig:10:1} Here we superimpose two red lines to the original distal stimulus (Hering illusion) to remark that vertical lines present in the stimulus are straight. \subref{fig:10:2} Representation of the displacement field $\{ \bar{u} (x_1, x_2) \}_{(x_1,x_2) \in \mathbb{R}^2}$. \subref{fig:10:3} Perceived deformation}
\end{figure*}

\subsection{From metric tensor field to image distortion}
\label{sec:25}

In the previous section we described the response of the cortex in the presence of a visual stimulus. 
\begin{itemize}
\item[(1)] The distal stimulus is projected onto the cortex by means of activity of simple cells. 
\end{itemize}
\begin{itemize}
\item[(2)]  The joint action of the short and long range connectivity 
induces a Riemannian tensor $\textbf{p}^{-1}$ on the $\mathbb{R}^2$ retinal plane. 
\end{itemize}
Even though it is not completely clear in which cortical area the perceived image is reconstructed, from a phenomenological point of view it is evident that our visual system recostructs the perceived image.

Hence a third mechanism takes place, 
able to construct the perceived stimulus from 
the cortical activation. With this mechanism the 
image distortion which induces 
the metric tensor $\textbf{p}$ (inverse of $\textbf{p}^{-1}$) is estimated. Here we 
propose to apply infinitesimal strain theory and to identify its inverse $\textbf{p}$ with the strain tensor to compute the deformation. 
Once the displacement vector field is applied to the distal stimulus, we obtain a distorted image which models the proximal one. In this way we justify the mechanism at the basis of geometrical optical illusions.

In this approach we consider the medium to be subjected only to small displacements, i.e. the geometry of the medium and its constitutive properties at each point of the space are assumed to be unchanged by deformation.

\subsubsection{Strain tensor - displacement vector field}
\label{sec:251}
The mathematical question is how to reconstruct the displacement starting from the strain tensor $\textbf{p}$. We think at the deformation induced by a geometrical optical illusion as a map between the $\mathbb{R}^2$ plane equipped with the metric $\textbf{p}$ and the $\mathbb{R}^2$ plane with the Euclidean metric $\textbf{Id}$: $$\Phi: (\mathbb{R}^2, \textbf{p}) \rightarrow  (\mathbb{R}^2, \textbf{Id}).$$
From the mathematical point of view this means that we look for the change of variable which induces the new metric, i.e. $$\bigg( \frac{\partial \Phi^k}{\partial x_i}\bigg) Id_{kl}\bigg(\frac{\partial \Phi^l}{\partial x_j}\bigg) = p_{ij}(x),$$ with $x =(x_1,x_2) \in \mathbb{R}^2$ (see Jost \cite{jost2008riemannian}), 
obtaining the relation: 
\begin{equation} \textbf{p}(x)=(\nabla \Phi)^T (\nabla \Phi). \label{RCGT}\end{equation}
Let us notice $\textbf{p}^{-1}$ corresponds to $\Phi^{-1}$, the map representing the process which builds the modulated connectivity we discussed before.
In strain theory $\textbf{p}$ satisfying \eqref{RCGT} is called \textit{right Cauchy-Green tensor} associated to the deformation $\Phi$, which from the physical point of view is a map $\Phi: \bar{\Omega} \rightarrow \mathbb{R}^2$ associating the points of the closure of a bounded open set $\Omega \subset \mathbb{R}^2$ (initial configuration of a body) to $\Phi (\Omega) \subset \mathbb{R}^2$ (deformed configuration). For references see \cite{lubliner2008plasticity}, \cite{marsden1994mathematical}.
It is possible to introduce the displacement as a map $\bar{u}(x_1,x_2) = \Phi(x_1,x_2)  - (x_1,x_2) $, where $(x_1,x_2) \in \mathbb{R}^2$. It follows $$\nabla \bar{u} = \nabla \Phi - Id.$$
We can now express the right Cauchy-Green tensor in terms of displacement: 
\begin{align*} \textbf{p} = p_{ij}(x) & =(\nabla \Phi)^T (\nabla \Phi) = (\nabla \bar{u} + Id)^T (\nabla \bar{u} + Id) \\ & = (\nabla \bar{u} )^T (\nabla \bar{u}) + (\nabla \bar{u} ) + (\nabla \bar{u})^T + Id .\end{align*}
The concept of strain is used to evaluate how much a given displacement differs locally from a rigid body displacement. One of the strain tensors for large deformations is the so called \textit{Green-Lagrangian strain tensor or Green-Saint Venant strain tensor} defined as: $$\mathit{E} = \frac{1}{2} (\textbf{p} - Id) = \frac{1}{2} ((\nabla \Phi)^T (\nabla \Phi) - Id), $$
which can be written in terms of the displacement as before: 
$$\mathit{E(\bar{u})} = \frac{1}{2} ((\nabla \bar{u} ) + (\nabla \bar{u})^T + (\nabla \bar{u} )^T (\nabla \bar{u})).$$ 
For \textit{infinitesimal deformations} of a continuum body, in which the displacement gradient is small ($\Arrowvert \nabla \bar{u}\Arrowvert \ll 1$), it is possible to perform a geometric linearization of strain tensors introduced before, in which the non-linear second order terms are neglected. The \textit{linearized Green-Saint Venant tensor} has the following form:
\begin{equation}
\mathit{E(\bar{u})} \approx \epsilon(\bar{u}) = \frac{1}{2}((\nabla \bar{u}) + (\nabla \bar{u})^T),
\end{equation}
which is used in the study of linearized elasticity, i.e. the study of such situations in which the displacements of the material particles of a body are assumed to be small (as stated at the beginning, infinitesimal strain theory.) \\
Here we give the expression in components of $\epsilon(\bar{u})$: 

\begin{equation}
\epsilon_{ij} (\bar{u}) =\left(\begin{array}{cc}
\frac{\partial {u}_1}{\partial x_1} &  \frac{1}{2} \big( \frac{\partial  {u}_1}{\partial x_2} + \frac{\partial  {u}_2}{\partial x_1} \big)\\
 \frac{1}{2} \big( \frac{\partial  {u}_2}{\partial x_1} + \frac{\partial  {u}_1}{\partial x_2} \big)&\frac{\partial {u}_2}{\partial x_2}  \\
  \end{array}\right),
\end{equation}
where $\bar{u} = (u_1,u_2)$.
Expressing $\epsilon_{ij}$ in terms of the metric $(p_{ij})_{i,j}$ with whom the initial configuration of the considered body was equipped we obtain:
\begin{equation}
\mathit{E} = \frac{1}{2} ((p_{ij})_{ij} - Id) \approx \epsilon_{ij} (\bar{u}),
\end{equation}
and in its matrix form: 
\begin{equation} \label{matrix form_GSV}
\left(\begin{array}{cc} p_{11} & p_{12} \\ p_{21} & p_{22}\\ \end{array}\right) - \left(\begin{array}{cc} 1 & 0 \\ 0 & 1\\ \end{array}\right)=\left(\begin{array}{cc}
\frac{\partial  {u}_1}{\partial x_1} &  \frac{1}{2} \big( \frac{\partial  {u}_1}{\partial x_2} + \frac{\partial  {u}_2}{\partial x_1} \big)\\
 \frac{1}{2} \big( \frac{\partial  {u}_2}{\partial x_1} + \frac{\partial  {u}_1}{\partial x_2} \big)&\frac{\partial  u_2}{\partial x_2}  \\
  \end{array}\right).
\end{equation}

\subsubsection{Poisson problems - displacement}
\label{sec:252}
Starting from \eqref{matrix form_GSV} we obtain a system of equations with this form:
\begin{eqnarray}
\left\{\begin{array}{ccc} p_{11} - 1 &=& \frac{\partial {u}_1}{\partial x_1}  \\ 
p_{22} - 1 &=& \frac{\partial u_2}{\partial x_2} \\
p_{12} &=& p_{21} = \frac{1}{2} (\frac{\partial}{\partial x_2}u_1 + \frac{\partial}{\partial x_1} u_{2}) \\
\end{array}\right.
\label{1eq}
\end{eqnarray}
Differentiating, substituting and imposing Neumann boundary conditions to system \eqref{1eq} we end up with the following differential system:
\begin{eqnarray} \label{laplac}
\left\{\begin{array}{cc} 
\Delta {u}_1 = \frac{\partial}{\partial x_1}p_{11} + 2\frac{\partial}{\partial x_2} p_{12} - \frac{\partial}{\partial x_1} p_{22} & \,\, \quad \mbox{in}\,\, M \\
\Delta {u}_2 = \, \frac{\partial}{\partial x_2} p_{22} + 2\frac{\partial}{\partial x_1}p_{12} -\frac{\partial}{\partial x_2} p_{11} &\\
& \\
\frac{\partial}{\partial \vec{n}} {u}_1 = 0 &\,\,\,\quad \mbox{in} \,\, \partial M  \\ \frac{\partial}{\partial \vec{n}} {u}_2 = 0 & \\
\end{array}\right. 
\end{eqnarray}
Let us explicitly note that tensor $\textbf{p}$ is obtained after convolution of Gabor filters, so that it is differentiable, allowing to write the system. Hence we solve \eqref{laplac}, recovering the displacement field $\bar{u} (x_1, x_2) $.

\section{Numerical implementation and results}\label{sec:3}
The inverse of tensor expressed in formula \eqref{ptensorvero} is computed discretizing $\theta$ as a vector of 32 values equally spaced in the interval $[0, \pi]$. The scale parameter $\sigma$ varies in dependence of the image resolution and is set in concordance with the stimulus processed. It is taken quite large in all examples in such a way to obtain a smooth tensor field covering all points of the image. This is in accordance with the hypothesis previously introduced that mechanisms in V2, where the receptive field of simple cells is larger than in V1, play a role in such phenomena. 
The constant $\gamma$ has been chosen for all the examples as $ \gamma= 2 \cdot 10^{-2} $.
The differential problem in \eqref{laplac} is approximated with a central finite difference scheme and it is solved with a classical PDE linear solver. We now start discussing all results obtained through the presented algorithm.

 \begin{figure}[H]
\begin{subfigure}[b]{1.6 in}
\centering
   \includegraphics[width= \columnwidth]{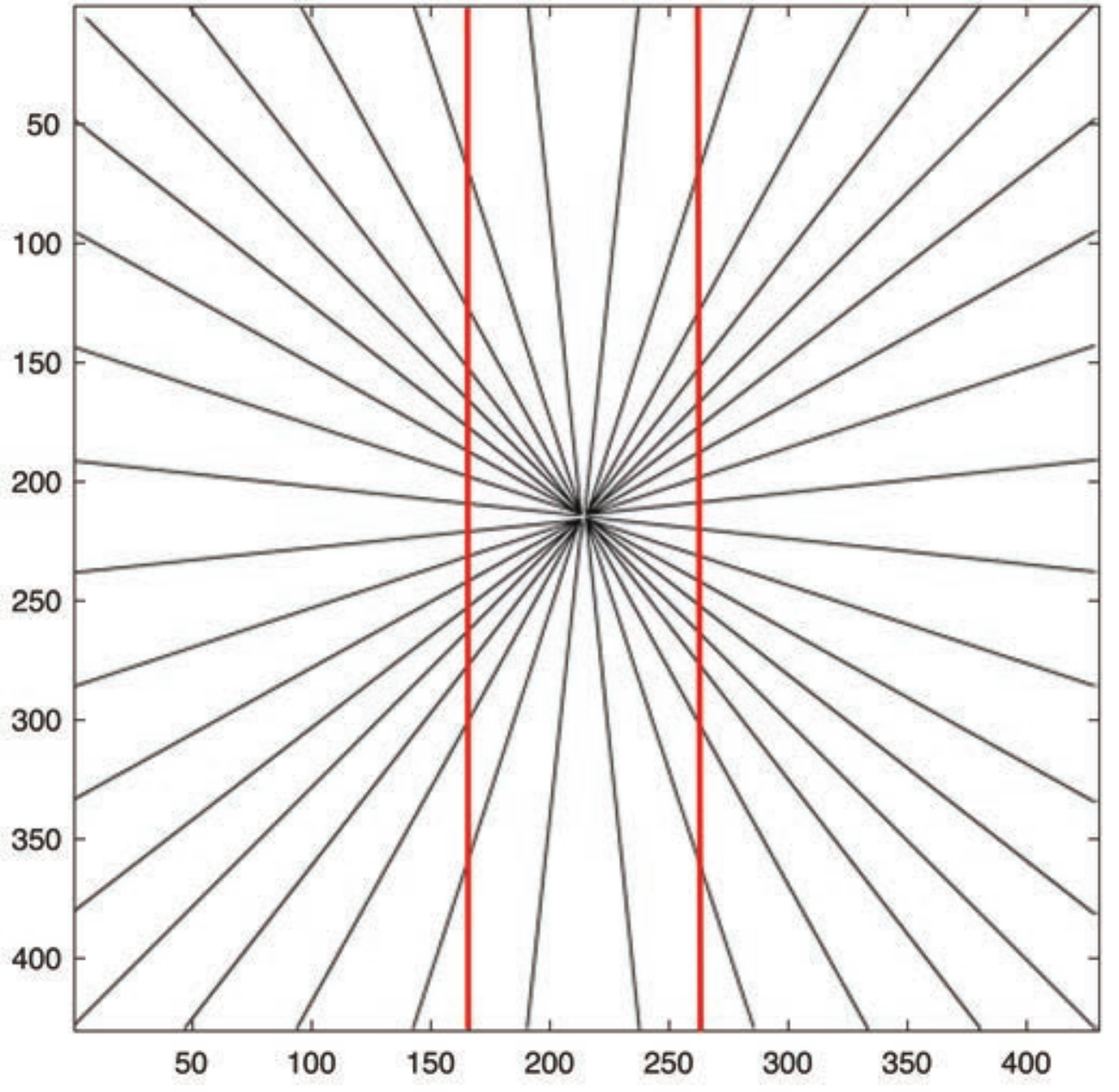}
          	\caption{}
       	\label{fig:11:1}

 	\end{subfigure}      	
    	\begin{subfigure}[b]{1.6 in}
    	\centering
    		\includegraphics[width= 1\columnwidth]{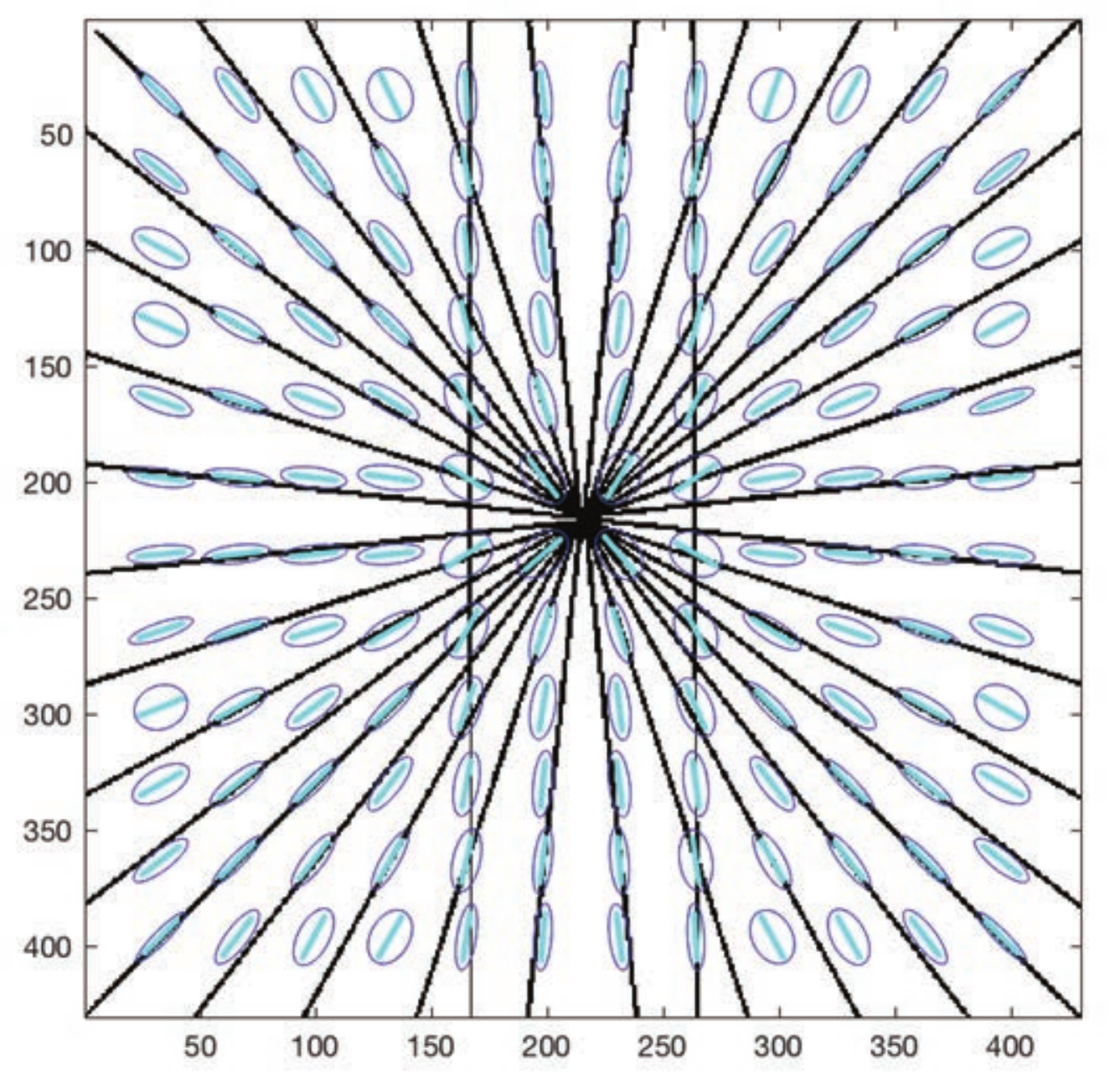} 
    		\caption{}
    		\label{fig:11:2}
    	\end{subfigure}
    	
\begin{subfigure}[b]{1.6 in}
\centering
    	    		\includegraphics[width= \columnwidth]{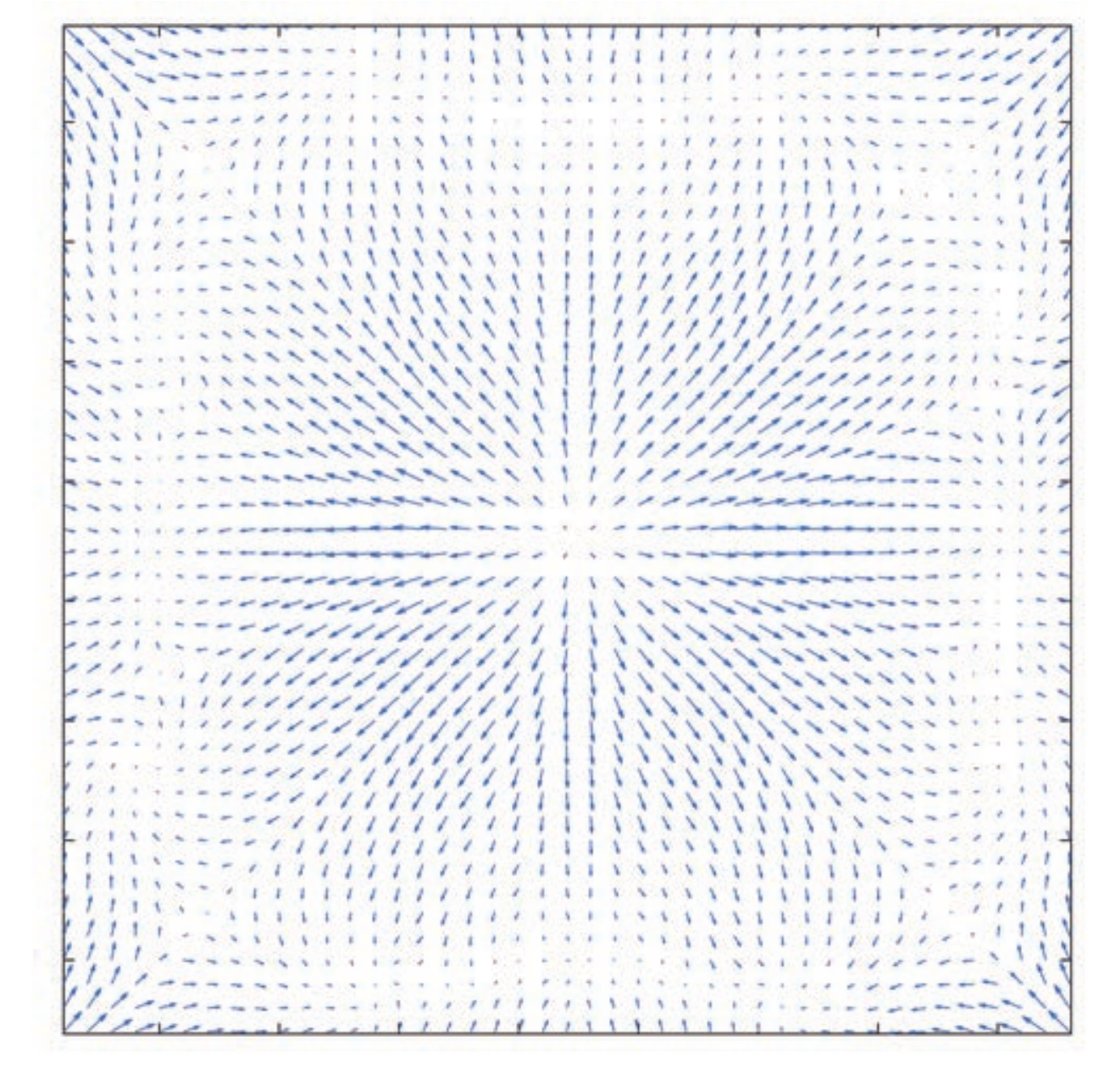} 
    	    		\caption{}
    	    	\label{fig:11:3}
\end{subfigure}
\begin{subfigure}[b]{1.6 in}
\centering
    	    		\includegraphics[width= \columnwidth]{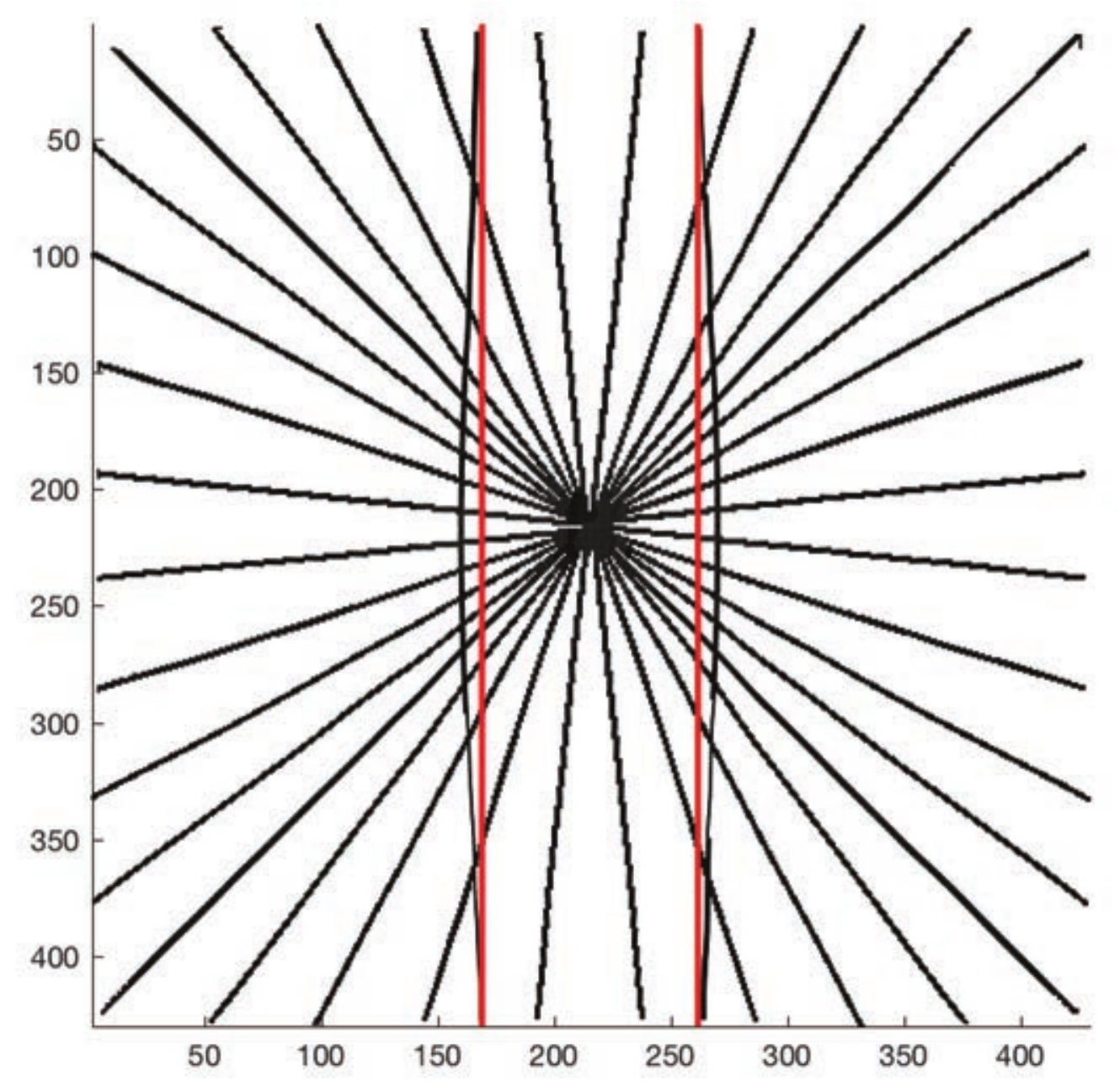} 
    	    		\caption{}
    	    	\label{fig:11:4}
\end{subfigure}

    	    	\caption{\subref{fig:11:1} We superimpose two red vertical lines to the Hering illusion, represented in figure \ref{fig:1:4}, in order to remark that vertical lines present in the stimulus are straight. \subref{fig:11:2} Representation of $\textbf{p}^{-1}$, projection onto the retinal plane of the polarized connectivity in \ref{ptensor}. The first eigenvalue has direction tangent to the level lines of the distal stimulus. In blue the tensor field, in cyan the eigenvector related to the first eigenvalue. \subref{fig:11:3} Computed displacement field $\bar{u}: \mathbb{R}^2 \rightarrow \mathbb{R}^2$. \subref{fig:11:4} Displacement applied to the image. In black we represent the proximal stimulus as displaced points of the distal stimulus: $(x_1, x_2) + \bar{u} (x_1, x_2)$. In red we give two straight lines as reference, in order to better clarify the curvature of the target lines}
    	    	 \label{her_5_fig1}
\end{figure}

\subsection{Hering illusion} \label{sec:31}
The Hering illusion, introduced by Hering, a German physiologist, in 1861 \cite{Her_1} is presented in figure \ref{fig:1:4}. In this illusion two vertical straight lines are presented in front of radial background, so that the lines appear as if they were bowed outwards. In order to help the reader, in figure \ref{fig:11:1} we superpose to the initial illusion two red vertical lines, which indeed coincide with the ones present in the stimulus. As described in the previous sections, we first convolve the distal stimulus with the entire bank of Gabor filters: we take 32 orientations selected in $[0,\pi)$, $\sigma = 6.72$ pixels. 
Following the process, we compute $\textbf{p}^{-1}$ using equation \eqref{ptensorvero}, we solve equation \eqref{laplac} obtaining the perceived displacement $\bar{u} : \mathbb{R}^2 \rightarrow \mathbb{R}^2$. Once it is applied to the initial stimulus, the proximal stimulus is recovered.
The result of computation is shown in figure \ref{her_5_fig1}. The distorted image folds the parallel lines (in black) against the straight lines (in red) of the original stimulus (figure \ref{fig:11:4}).

\begin{figure}[H]
\centering 
\includegraphics[width=\columnwidth]{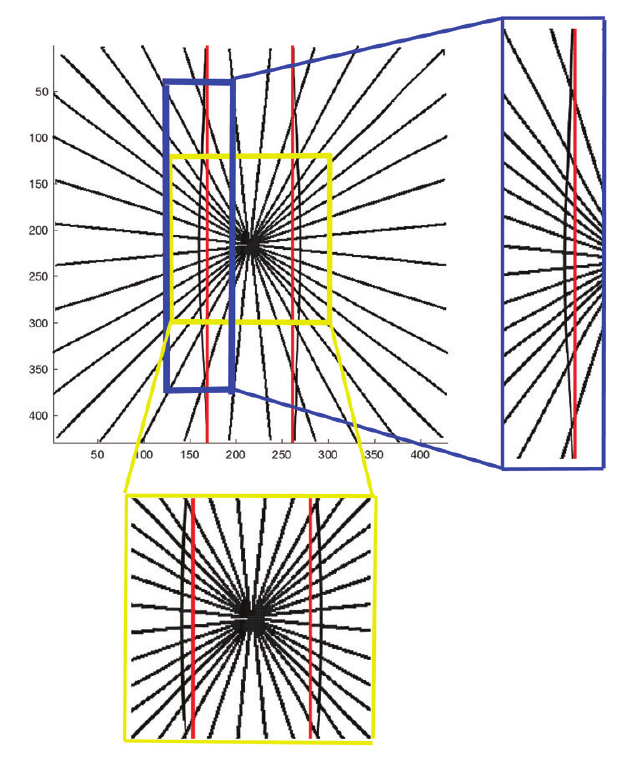}
\caption{Details of the perceived distortion in the computed proximal stimulus}
\label{fig:14}
\end{figure}

\subsection{Wundt Illusion} \label{sec:32}
\begin{figure}[H]

\centering
\begin{subfigure}[b]{1.6 in}
\centering
\includegraphics[width=\columnwidth]{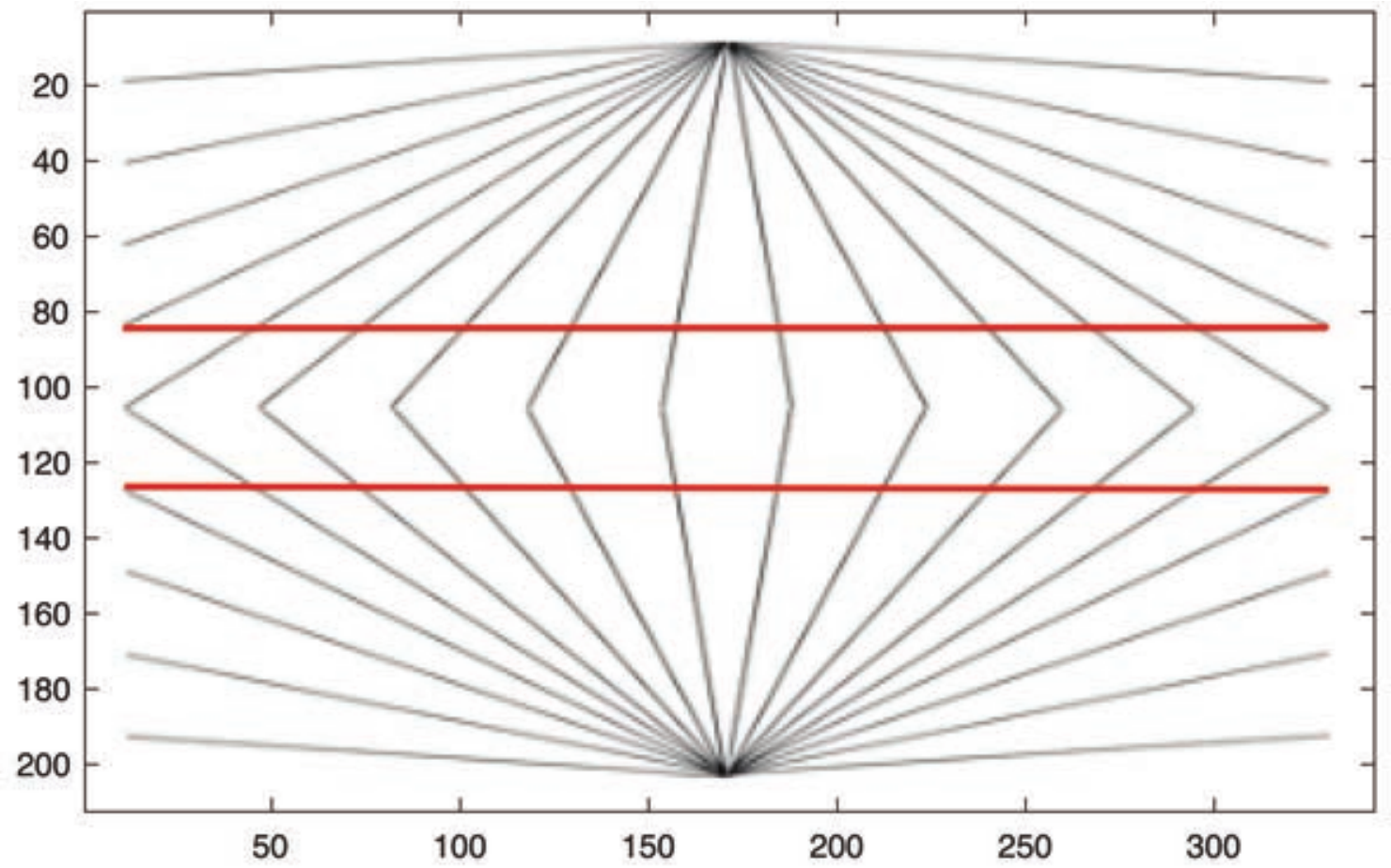}
\caption{}
\label{fig:15:1}
\end{subfigure}
\begin{subfigure}[b]{1.6 in}
\centering
\includegraphics[width=\columnwidth]{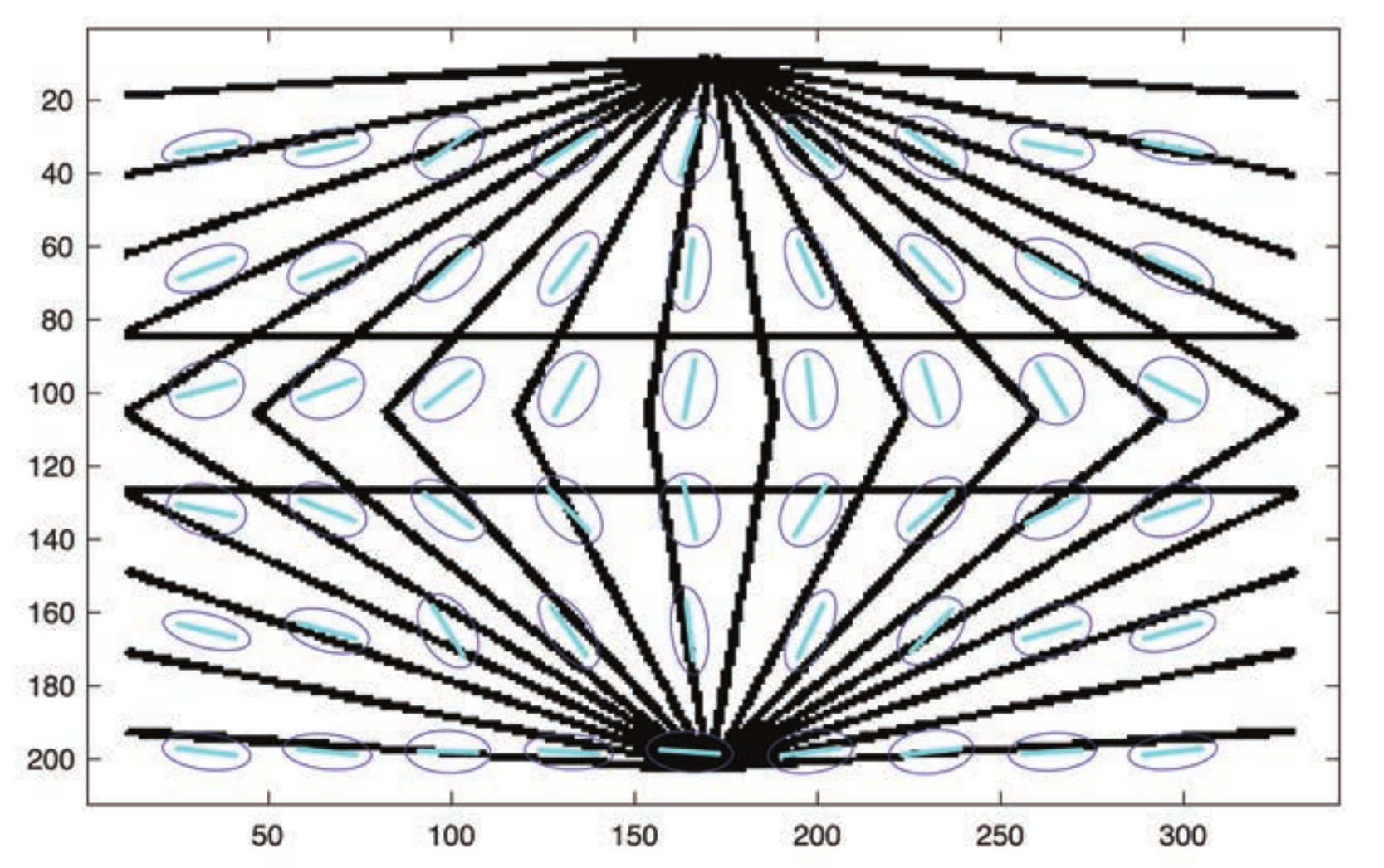}
\caption{}
\label{fig:15:2}
\end{subfigure}
\begin{subfigure}[b]{1.6 in}
\centering
\includegraphics[width=\columnwidth]{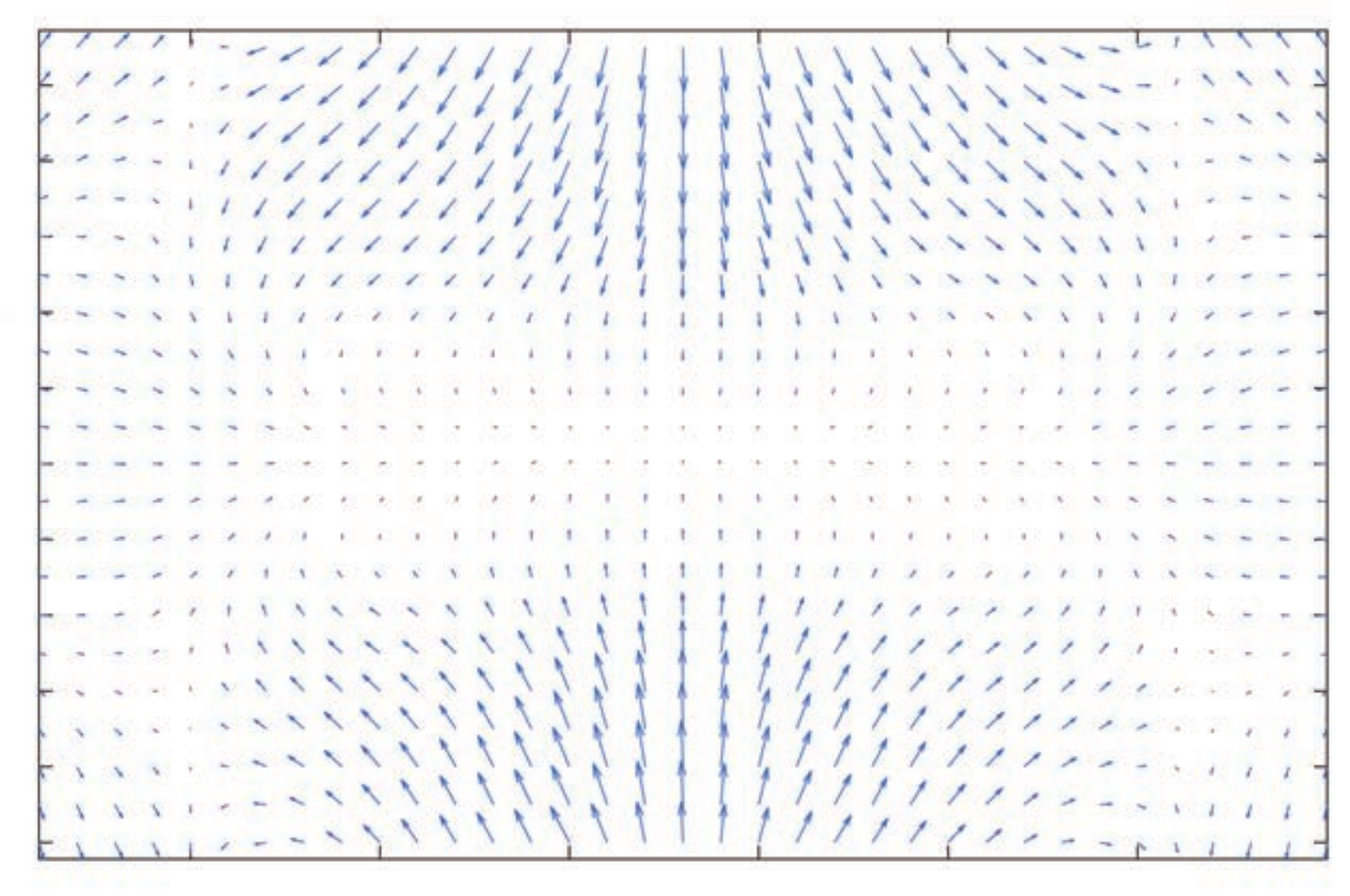}
\caption{}
\label{fig:15:3}
\end{subfigure}
\begin{subfigure}[b]{1.6 in}
\centering
\includegraphics[width=\columnwidth]{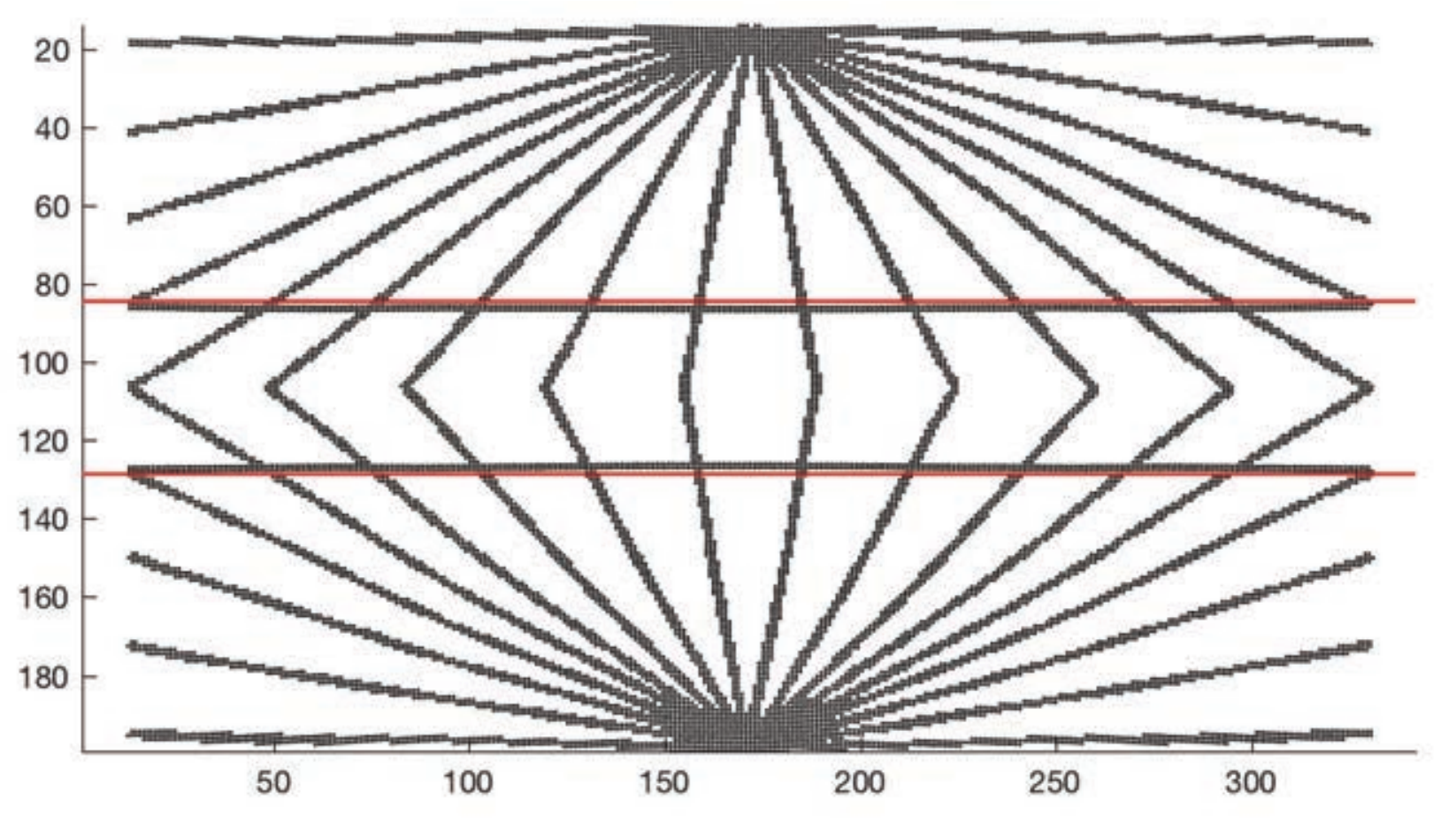}
\caption{}
\label{fig:15:4}
\end{subfigure}
\caption{\subref{fig:15:1} Here we superimpose two red lines to the Wundt illusion, 
presented in figure  \ref{fig:1:5}, 
in order to clarify that the horizontal lines present in the image are indeed straight. \subref{fig:15:2} Representation of $\textbf{p}^{-1}$, projection onto the retinal plane of the polarized connectivity in \ref{ptensor}. The first eigenvalue has direction tangent to the level lines of the distal stimulus. In blue the tensor field, in cyan the eigenvector related to the first eigenvalue. \subref{fig:15:3} Computed displacement field $\bar{u}$. \subref{fig:15:4} Displacement applied to the image. In black we represent the proximal stimulus as displaced points of the distal stimulus: $(x_1, x_2) + \bar{u} (x_1, x_2)$. In red we give two straight lines as reference, in order to put in evidence the curvature of the target lines}
\label{wundt_sec5}
\end{figure}

\begin{figure}[H]
\centering
\includegraphics[width=\columnwidth]{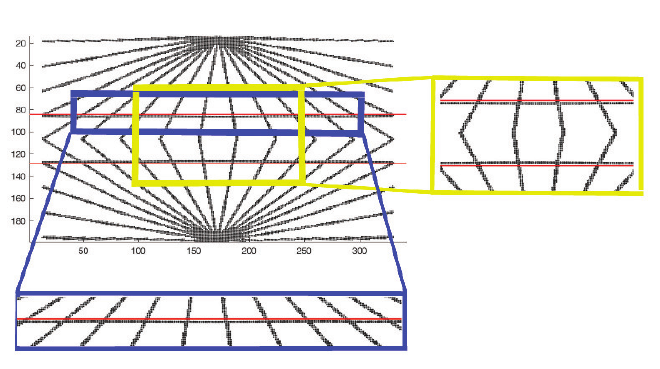}
\caption{Details of the perceived distortion in the computed proximal stimulus}
\label{fig:18}
\end{figure}

A variant of the Hering illusion, introduced by Wundt in the 19th century, \cite{wundt1898geometrisch} is presented in figure \ref{fig:1:5}. In this illusion two straight horizonal lines look as if they were bowed inwards, due to the distortion induced by the crooked lines on the background. 
For the convolution of the distal stimulus with Gabor filters we select 32 orientations in $[0,\pi)$, $\sigma = 11.2$ pixels. Then we apply the previous model, and obtain the result presented in figure \ref{wundt_sec5}.
Computed vector fields are concentrated in the central part of the image and point toward the center. They indicate the direction of the displacement, which bends the parallel lines inwards. In figure \ref{fig:15:4} the proximal stimulus is computed through the expression: $(x_1, x_2) + \bar{u} (x_1, x_2)$. In black we indicate displaced dots of the initial image: the straight lines of the distal stimulus are bent by the described mechanism (black). In red we put the straight lines of the original distal stimulus. This provide a comparison between the lines pre/post processing. In figure \ref{fig:18} details of the distances between the bent curves and the original straight lines are shown.

\subsection{Square shape over Ehrenstein context} \label{sec:33}
\begin{figure}[H]

\begin{subfigure}[b]{1.6 in}
\centering
\includegraphics[width=\columnwidth]{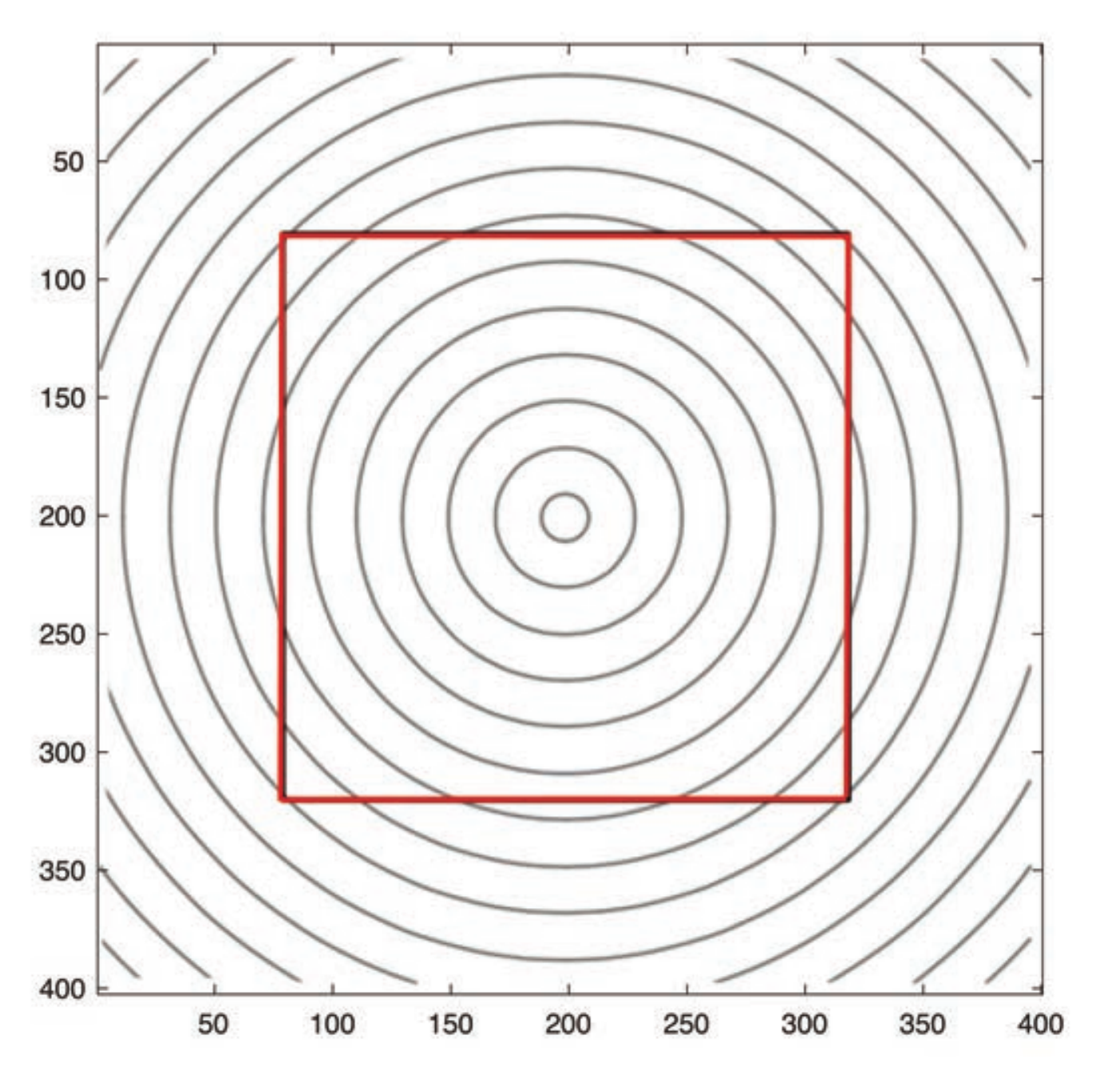}
\caption{}
\label{fig:19:1}
\end{subfigure}
\begin{subfigure}[b]{1.6 in}
\centering
\includegraphics[width=\columnwidth]{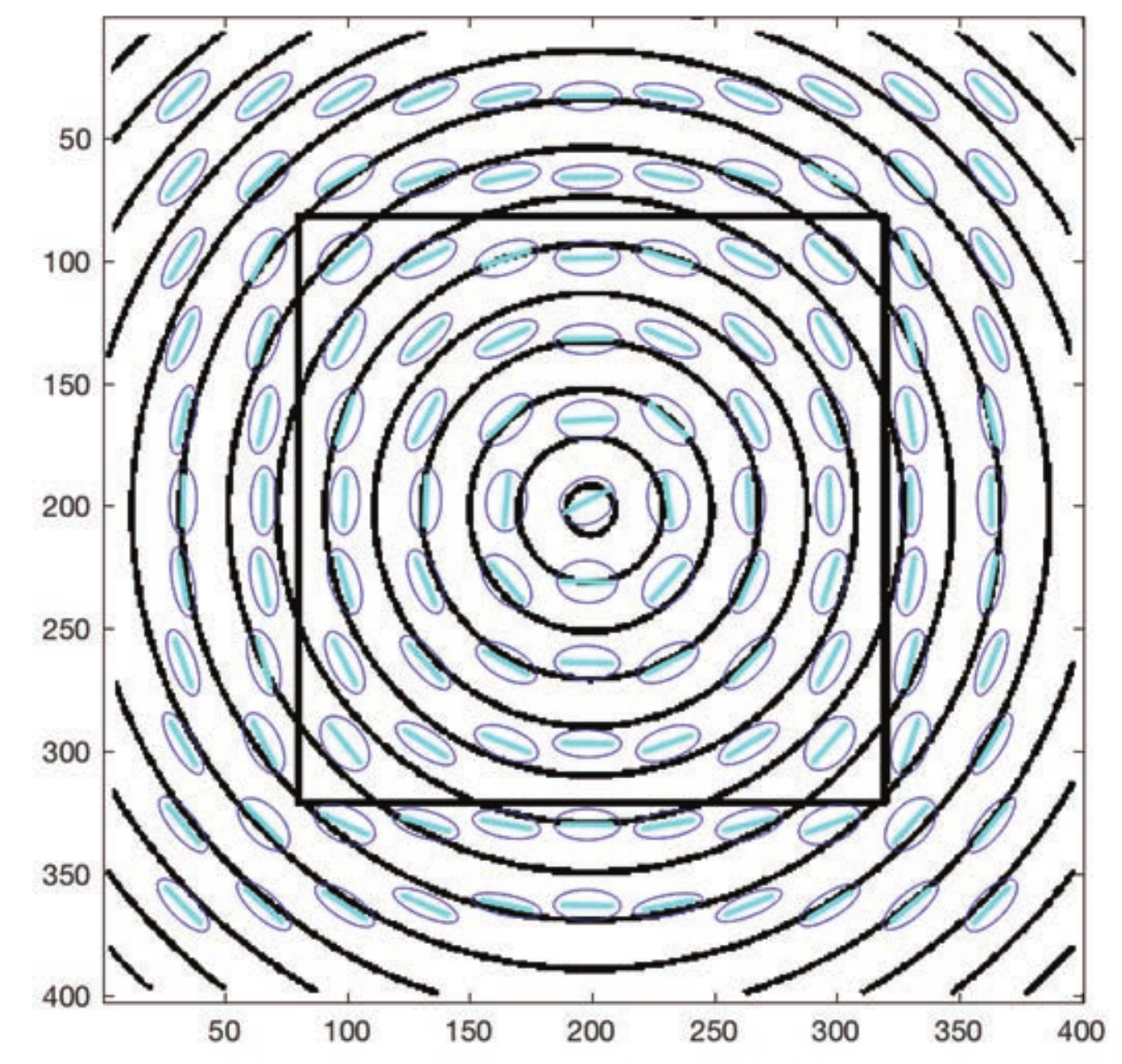}
\caption{}
\label{fig:19:2}
\end{subfigure}
\begin{subfigure}[b]{1.6 in}
\centering
\includegraphics[width=\columnwidth]{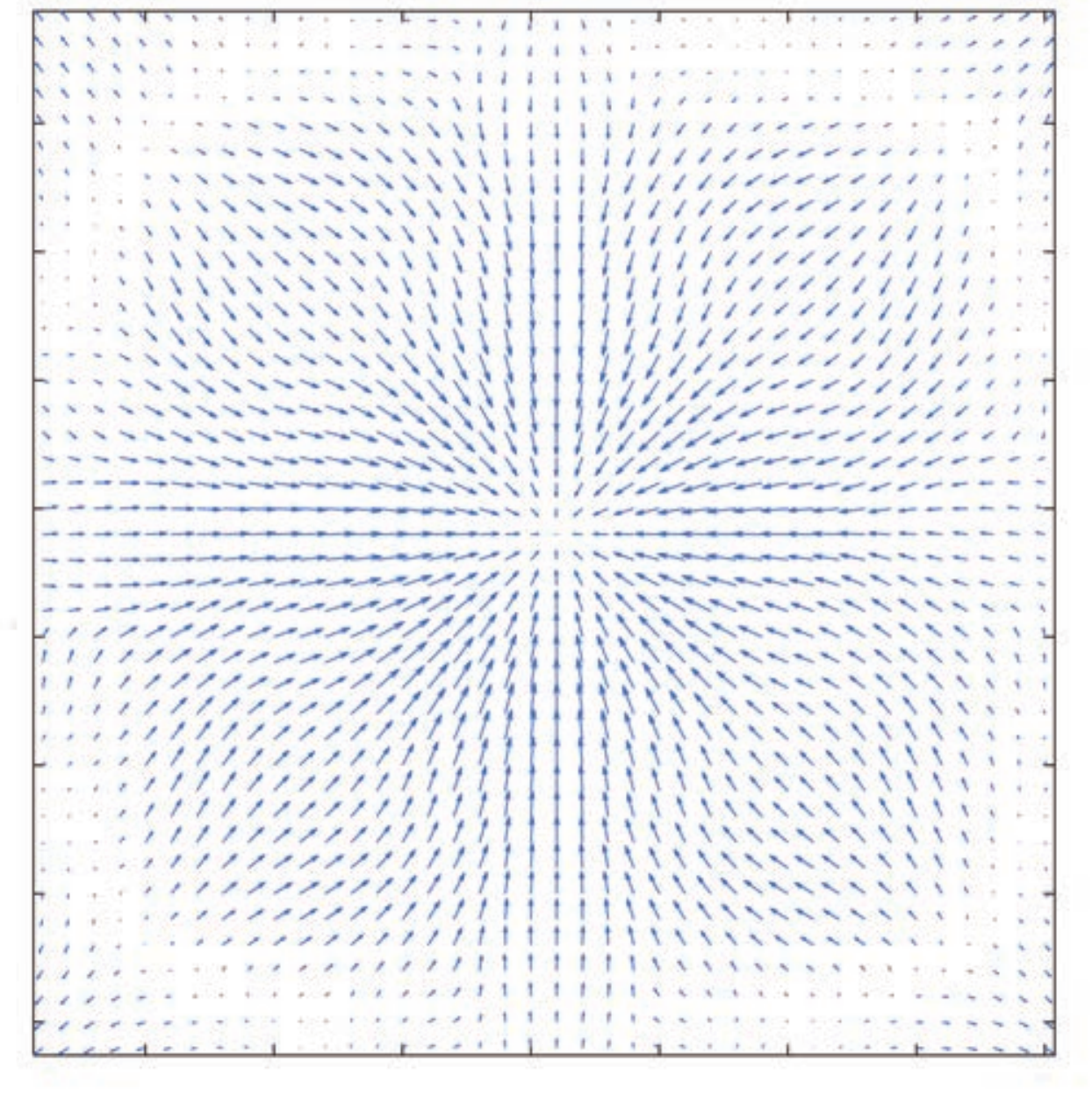}
\caption{}
\label{fig:19:3}
\end{subfigure}
\begin{subfigure}[b]{1.6 in}
\centering
\includegraphics[width=1.05\columnwidth]{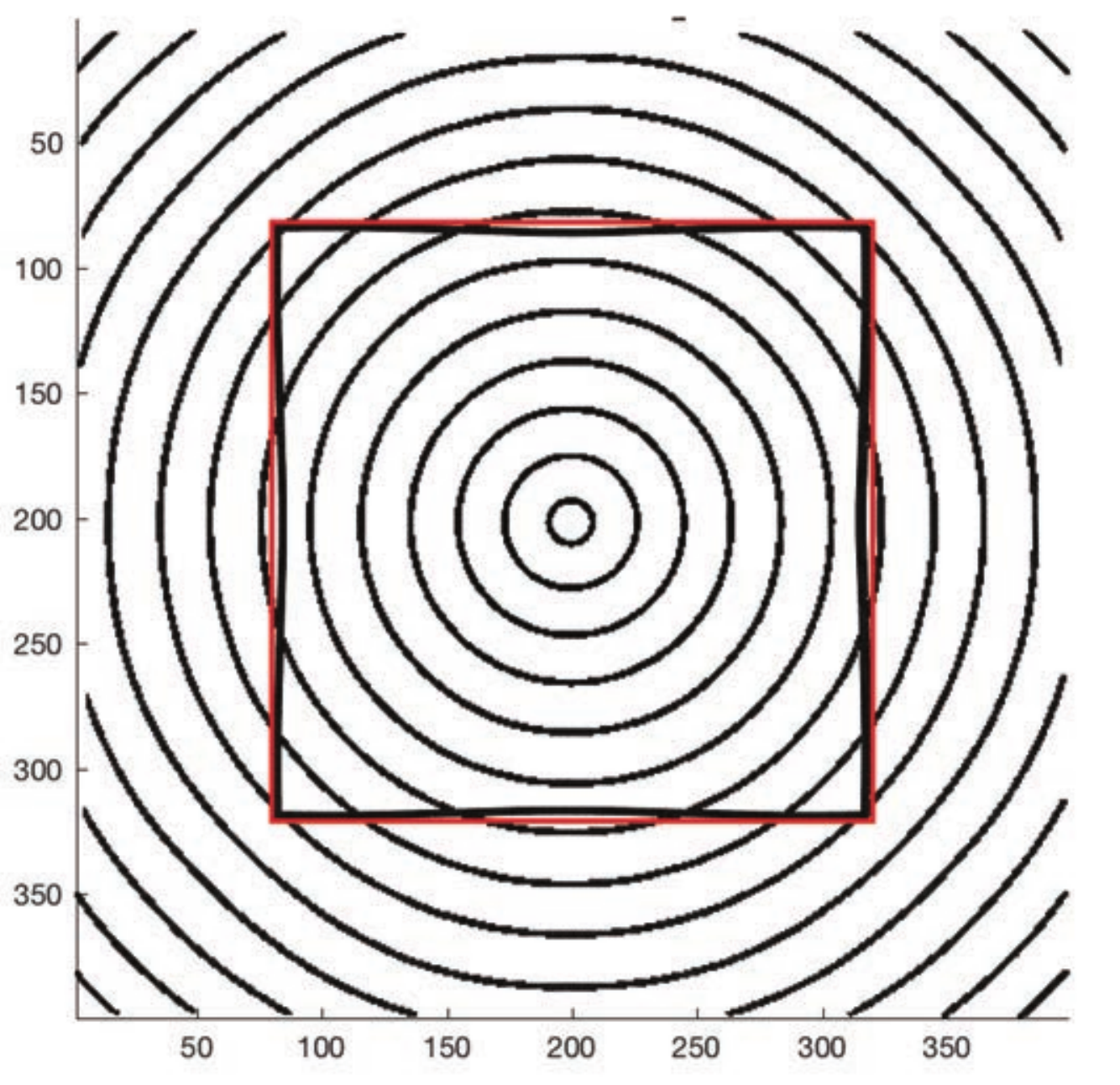}
\caption{}
\label{fig:19:4}
\end{subfigure}
\caption{\subref{fig:19:1} Here we superimpose red edges to the original illusion shown in \ref{fig:1:1}. \subref{fig:19:2} Representation of $\textbf{p}^{-1}$, projection onto the retinal plane of the polarized connectivity in \ref{ptensor}. The first eigenvalue has direction tangent to the level lines of the distal stimulus. In blue the tensor field, in cyan the eigenvector related to the first eigenvalue. \subref{fig:19:3} Computed displacement field $\bar{u}$. \subref{fig:19:4} Displacement applied to the image. In black we represent the proximal stimulus as displaced points of the distal stimulus: $(x_1, x_2) + \bar{u} (x_1, x_2)$. In red we give a square as reference, in order to put in evidence the curvature of the target lines}
\end{figure}

This illusion, introduced by Ehm and Wackermann in \cite{ehm2012modeling}, consists in presenting a square over a background of concentric circles, figure \ref{fig:1:1}. This context, the same we find in Ehrenstein illusion, bends the edges of the square (red lines in \ref{fig:19:1}) toward the center of the image. Here we take the same number of orientations, 32, selected in $[0,\pi)$ and $\sigma = 13.44$ pixels. The resulting distortion is shown in figure \ref{fig:19:4}.

\subsection{Modified Hering illusion} \label{sec:34}
\begin{figure}[htbp]
\begin{subfigure}{1.6 in}
\centering
 \includegraphics[width= 0.95\columnwidth]{Fig1a-eps-converted-to.pdf}
   \caption{}
  \label{fig:22:1}
  \end{subfigure}
   	\begin{subfigure}{1.6 in}
   		\includegraphics[width= \columnwidth]{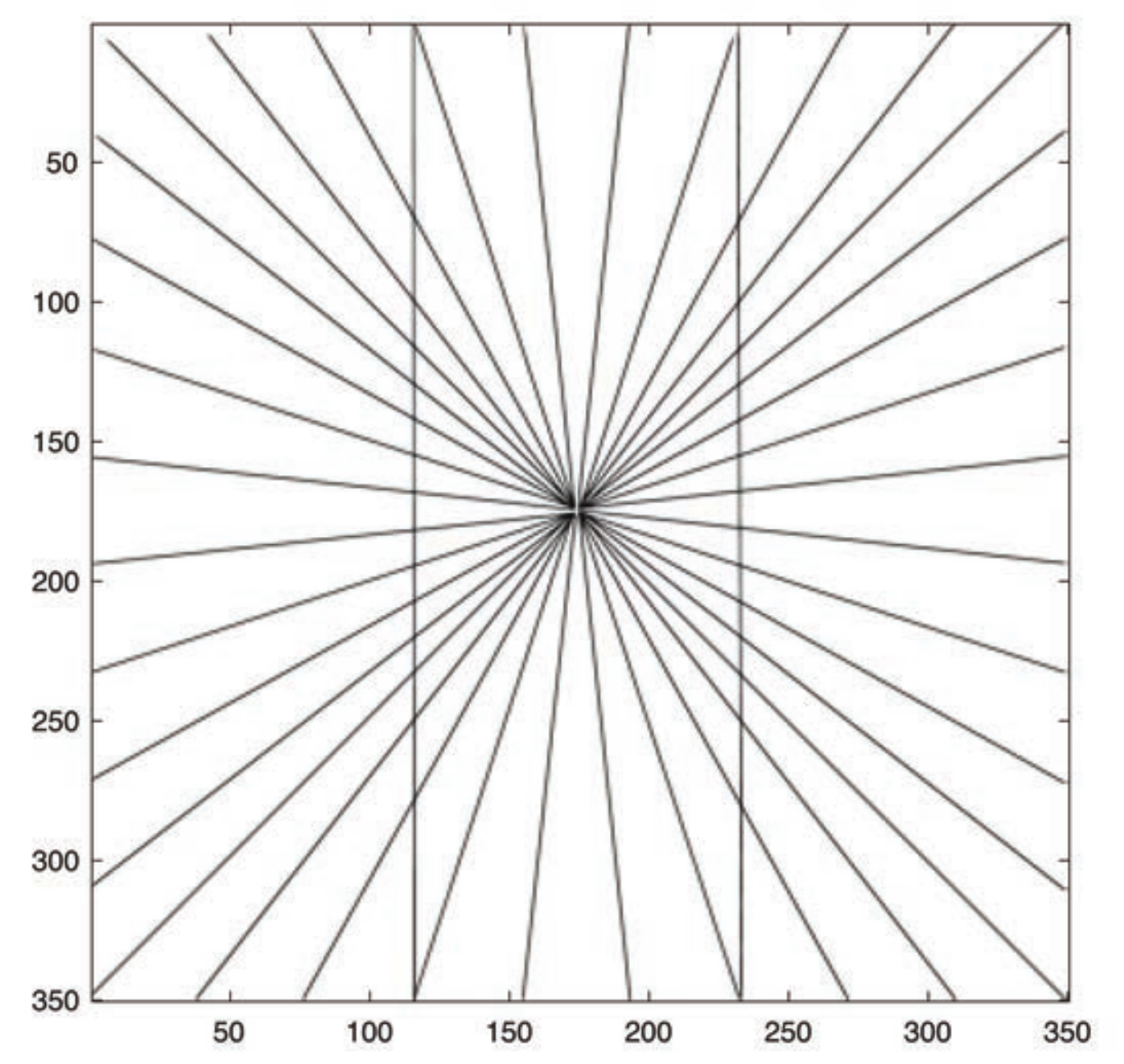} \caption{}
   		\label{fig:22:2}
   	\end{subfigure}
   	 	\begin{subfigure}{1.6 in}		
   	 		\includegraphics[width= \columnwidth]{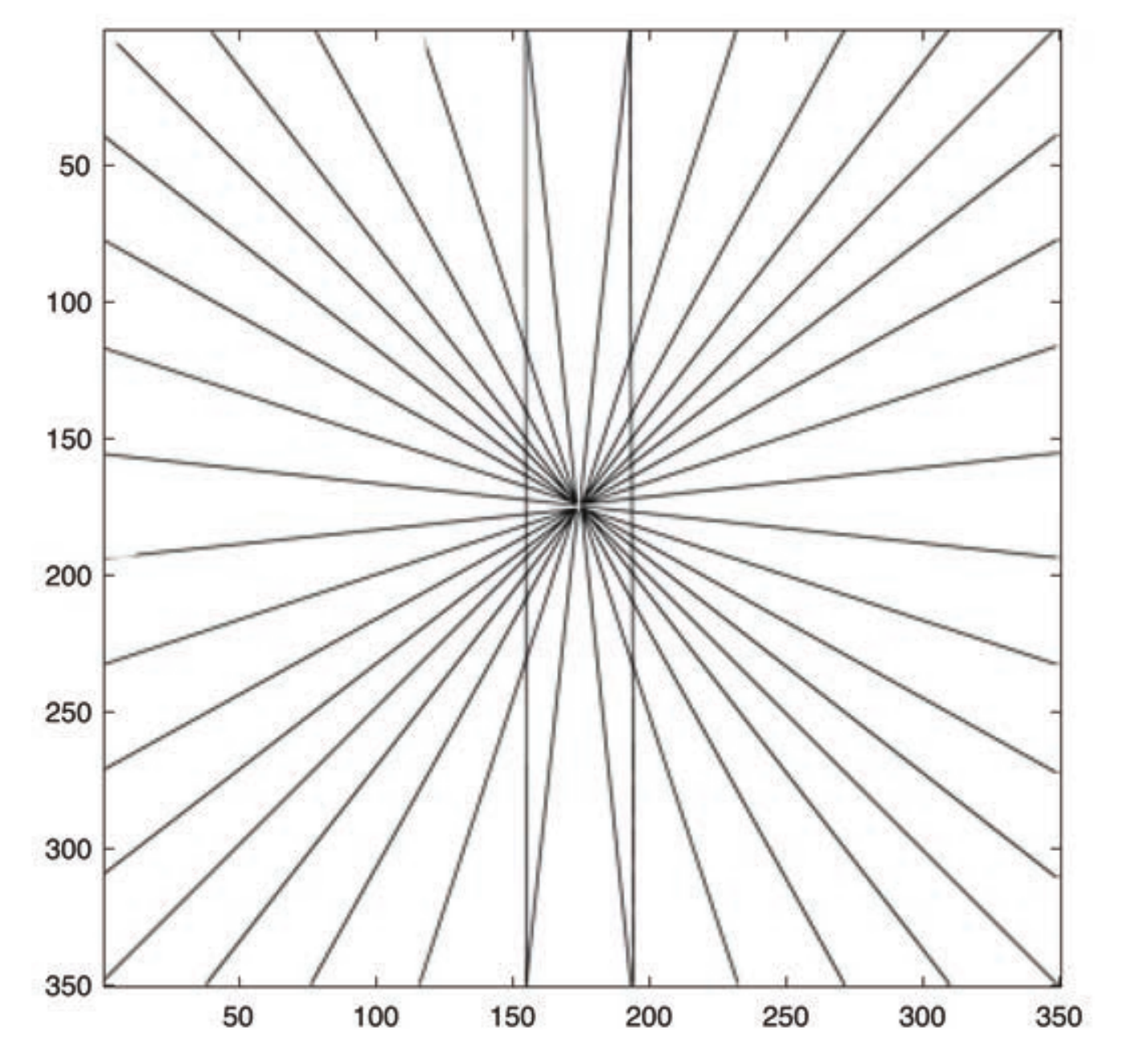}  \caption{}
   	 		\label{fig:22:3} 
   	 	\end{subfigure}
\begin{subfigure}{1.6 in}		
   	 	   	 		\includegraphics[width= \columnwidth]{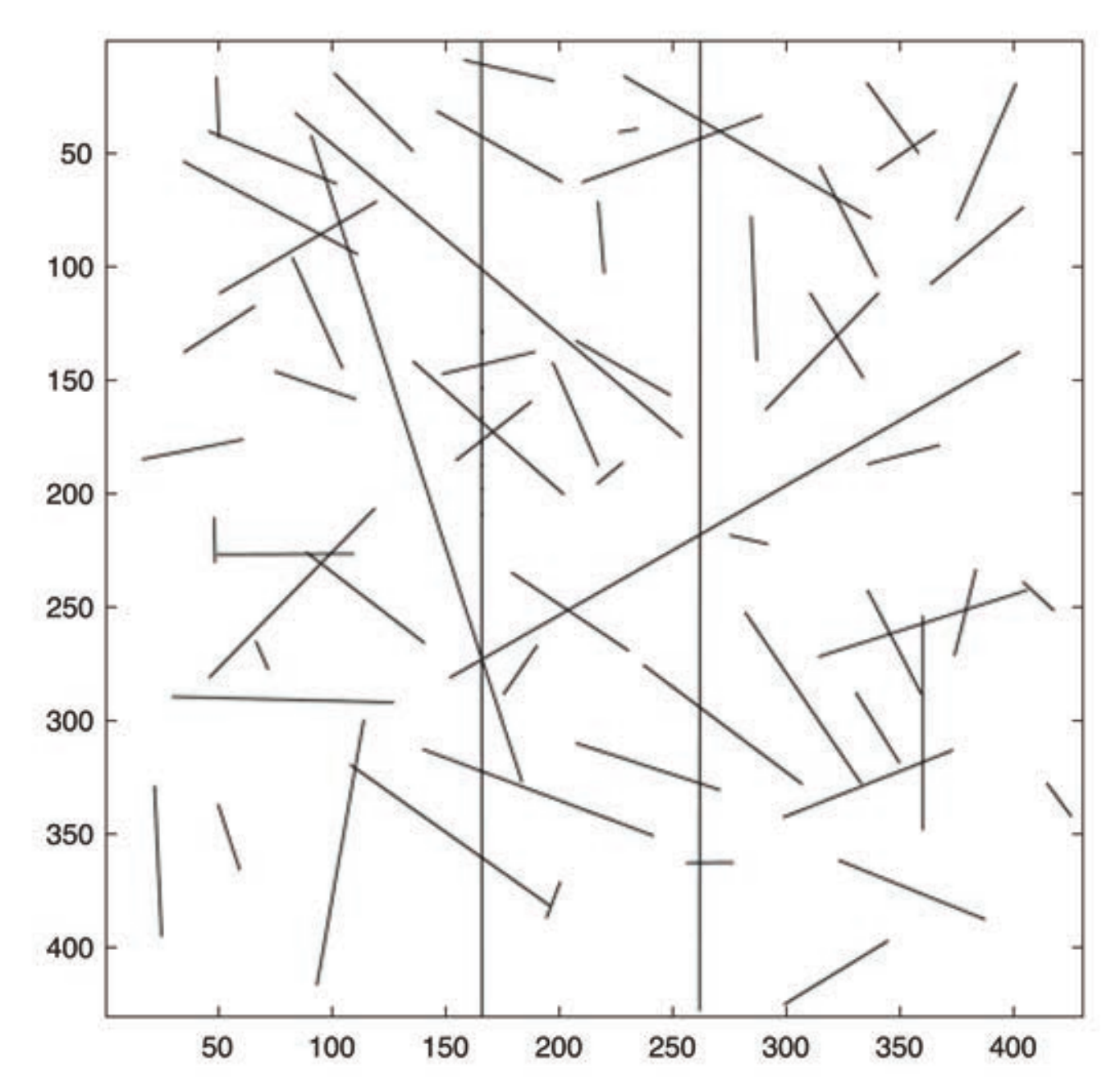}  \caption{}
   	 	   	 		\label{fig:22:4} 
\end{subfigure}
 	\caption{
 		\subref{fig:22:1} Hering illusion, distal stimulus. \subref{fig:22:2} Modified Hering illusion, in this example straight lines are further from the center with respect to the classical example of Hering illusion. \subref{fig:22:3} Modified Hering illusion, in this example straight lines are placed nearer the center with respect to the classical example of the Hering illusion. \subref{fig:22:4} Modified Hering illusion with a incoherent background, composed by random-oriented segments}
 		\label{mod_her}
 \end{figure}

Here we present three modified Hering illusions (see figure \ref{mod_her}): in the first one straight lines are positioned further from the center than in the classical Hering illusion. In the second one straight lines are positioned nearer the center than in the reference Hering illusion. For coherence with the Hering example, orientations selected are 32 in $[0,\pi)$ and $\sigma = 6.72$ pixels. All other parameters are fixed during these three experiments.  

\begin{figure}[H]
\begin{subfigure}{1.6 in}
\centering
 \includegraphics[width= \columnwidth]{Her1-eps-converted-to.pdf}
  \caption{}
  \label{fig:23:1}

  \end{subfigure}
   	\begin{subfigure}{1.6 in}
   		\includegraphics[width= \columnwidth]{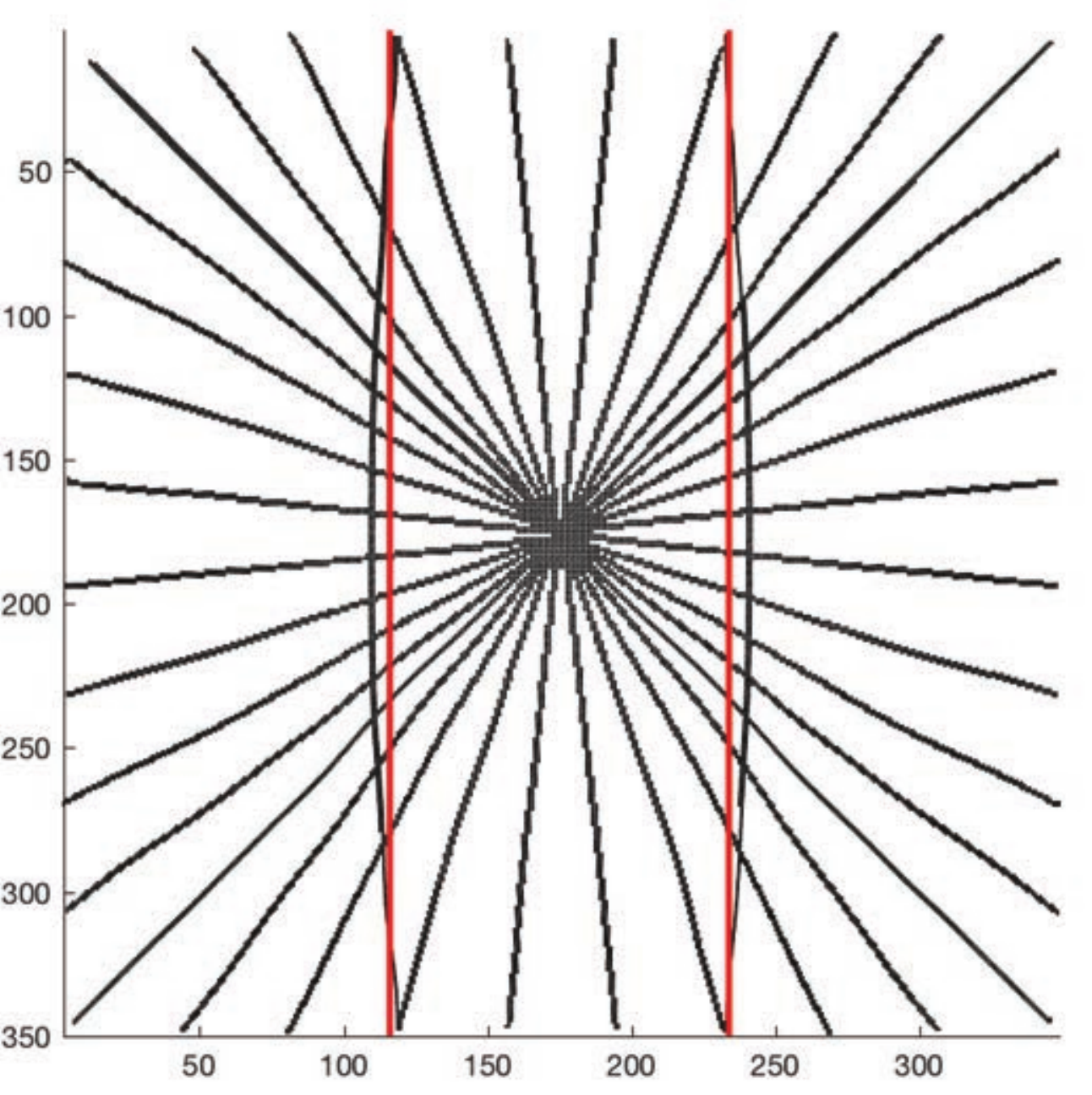} \caption{}
   		\label{fig:23:2}
   	\end{subfigure}
   	 	\begin{subfigure}{1.6 in}		
   	 		\includegraphics[width= 1.05\columnwidth]{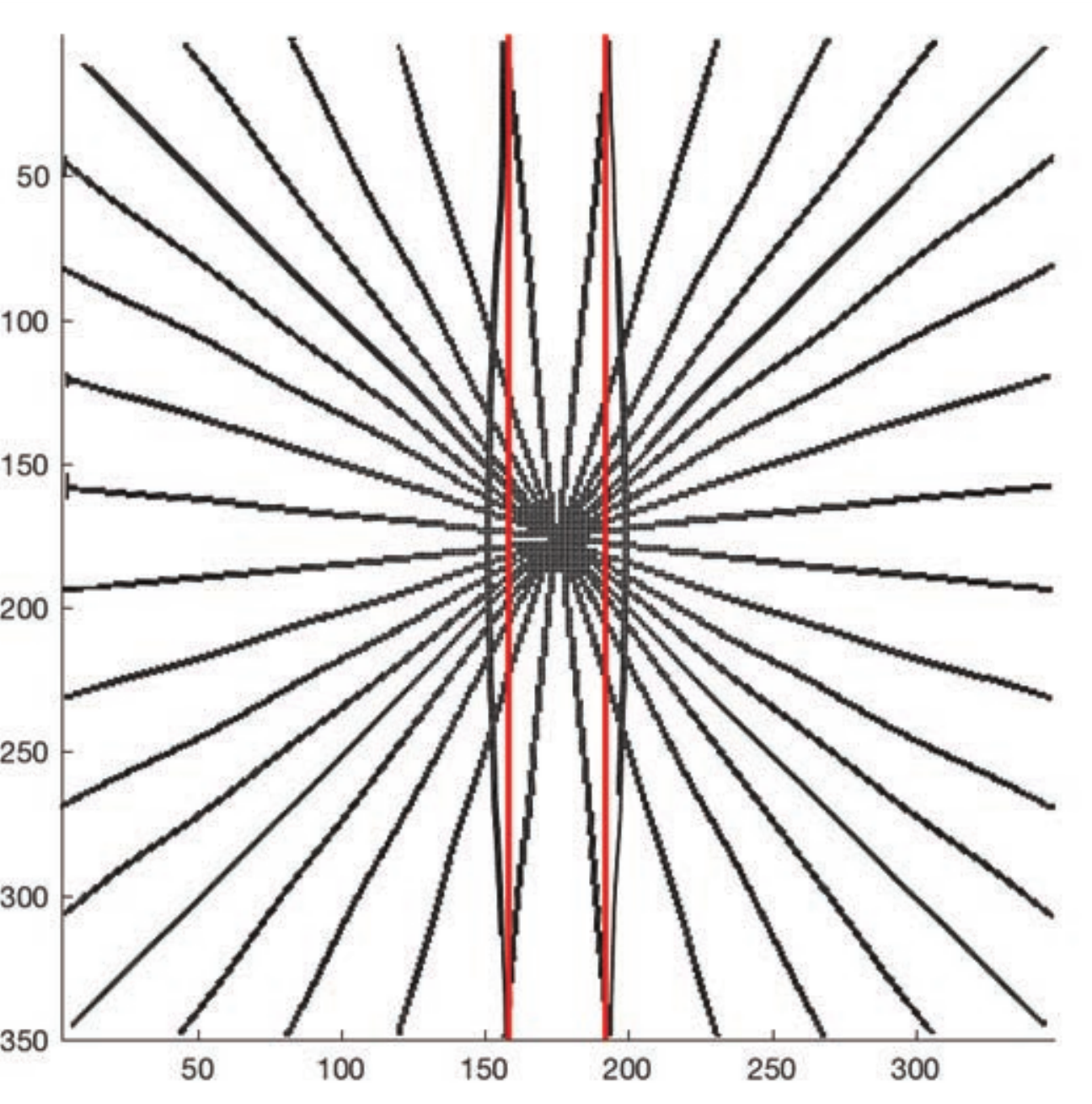}  \caption{}
   	 		\label{fig:23:3} 
   	 	\end{subfigure}
\begin{subfigure}{1.6 in}		
   	 	   	 		\includegraphics[width= \columnwidth]{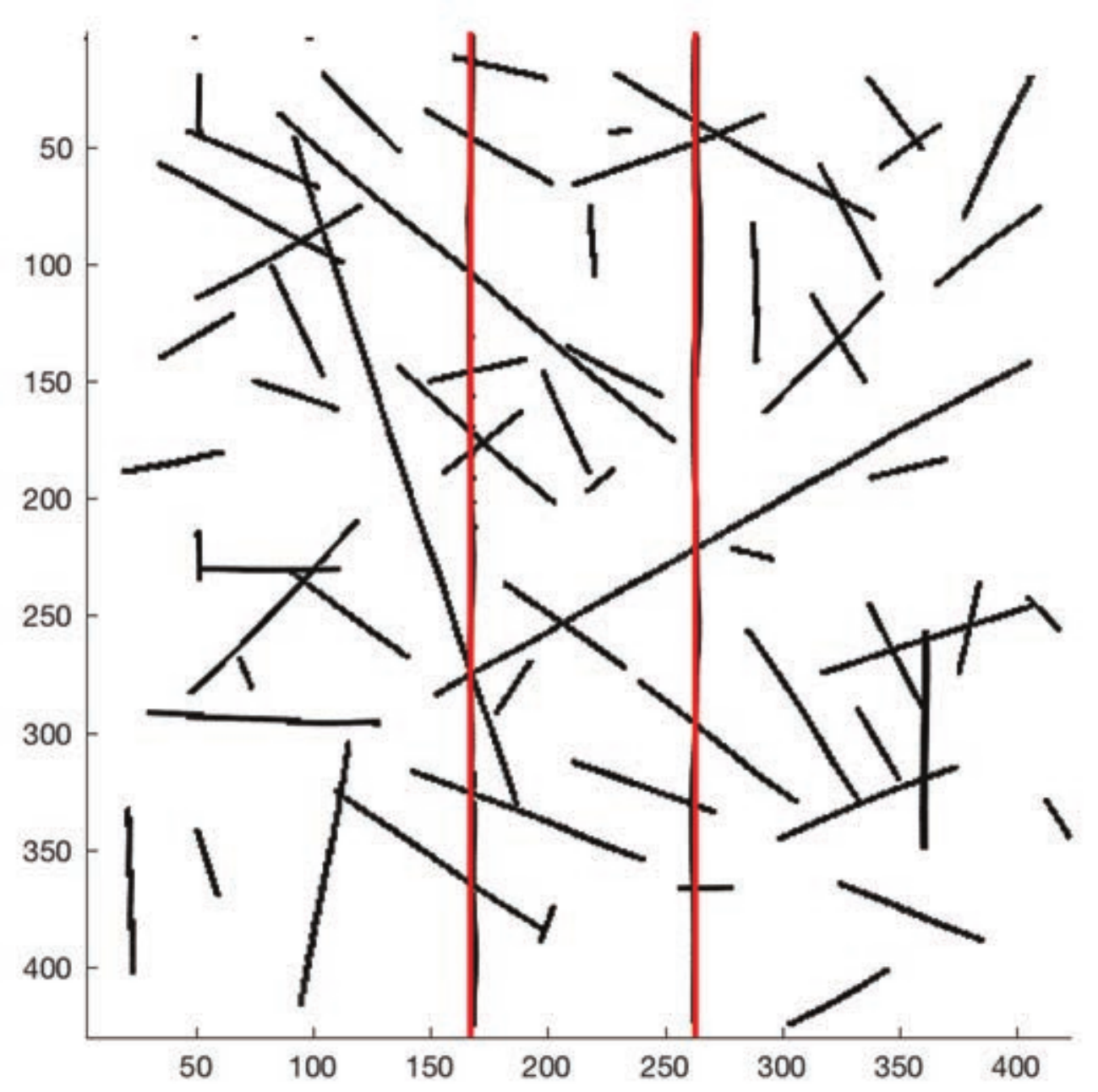}  \caption{}
   	 	   	 		\label{fig:23:4} 
\end{subfigure}
 	\caption{
 		\subref{fig:23:1} Displacement applied to the Hering illusion. \subref{fig:23:2} Displacement applied to the first modified Hering illusion, in which the distance from the center is increased. \subref{fig:23:3} Displacement applied to second modified Hering illusion, in which the distance from the center is decreased.  \subref{fig:23:4} Displacement applied to the third modified Hering illusion, with an incoherent background of random-oriented segments. In this last example no deformation is perceived. In black we represent the proximal stimulus as displaced points of the distal stimulus: $(x_1, x_2) + \bar{u} (x_1, x_2)$. In red we give two straight lines as reference, in order to put in evidence how much target lines are bent \subref{fig:23:1}, \subref{fig:23:2}, \subref{fig:23:3} or not bent \subref{fig:23:4}}
 \end{figure}

 In the proposed modified Hering illusions the vertical lines are straight and parallel as in the Hering, but since they are located further/nearer the center of the image the perceived bending results to be less/more intense. 
In accordance with the displacement vector fields shown in figure \ref{fig:11:3}, as far as we outstrip/approach the center the magnitude of the computed displacement decreases/increases. In figure \ref{fig:22:4} two straight lines are put over an incoherent background, composed by random oriented segments. As we can see from \ref{fig:23:4}, any displacement is perceived 
nor computed by the present algorithm.
\subsection{Wundt-Hering illusion} \label{sec:35}
The Wundt-Hering illusion (figure \ref{fig:1:6}) combines the effect of the background of the Hering and Wundt illusions. In this illusion two straight horizontal lines are presented in front of inducers which bow them outwards and inwards at the same time, inhibiting the bending effect. As a consequence the horizontal lines are indeed perceived as straight. As previously explained for the modified Hering illusion, also this phenomenon can be interpreted in terms of lateral interaction between cells belonging to the same neighborhood. Here we take 32 orientations selected in the interval $[0,\pi)$, $\sigma = 6.72$ pixels. 

\begin{figure}[H]
\centering
\begin{subfigure}[b]{1.6 in}
\includegraphics[width=\columnwidth]{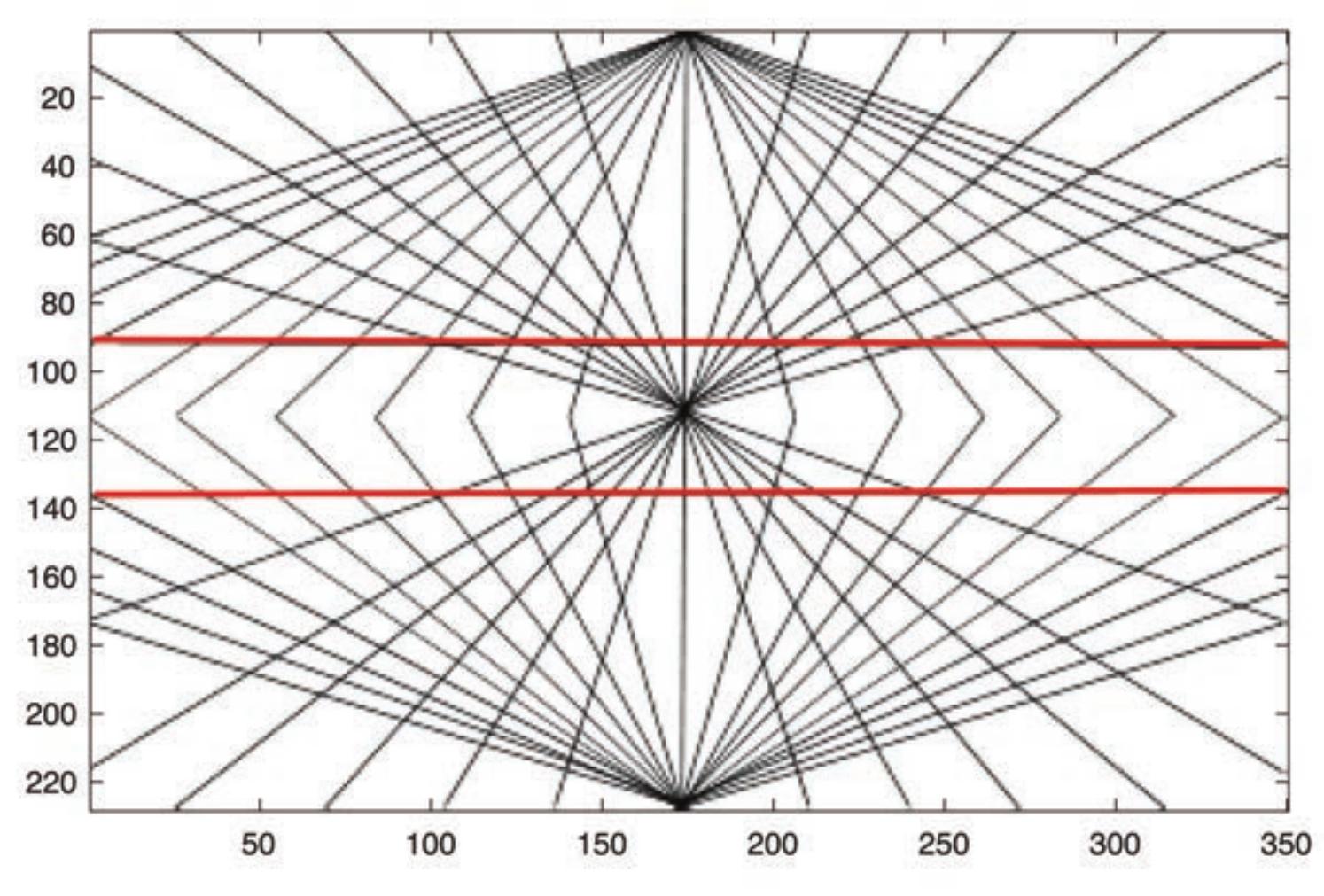}
\caption{}
\label{fig:24:1}
\end{subfigure}
\begin{subfigure}[b]{1.6 in}
\includegraphics[width=\columnwidth]{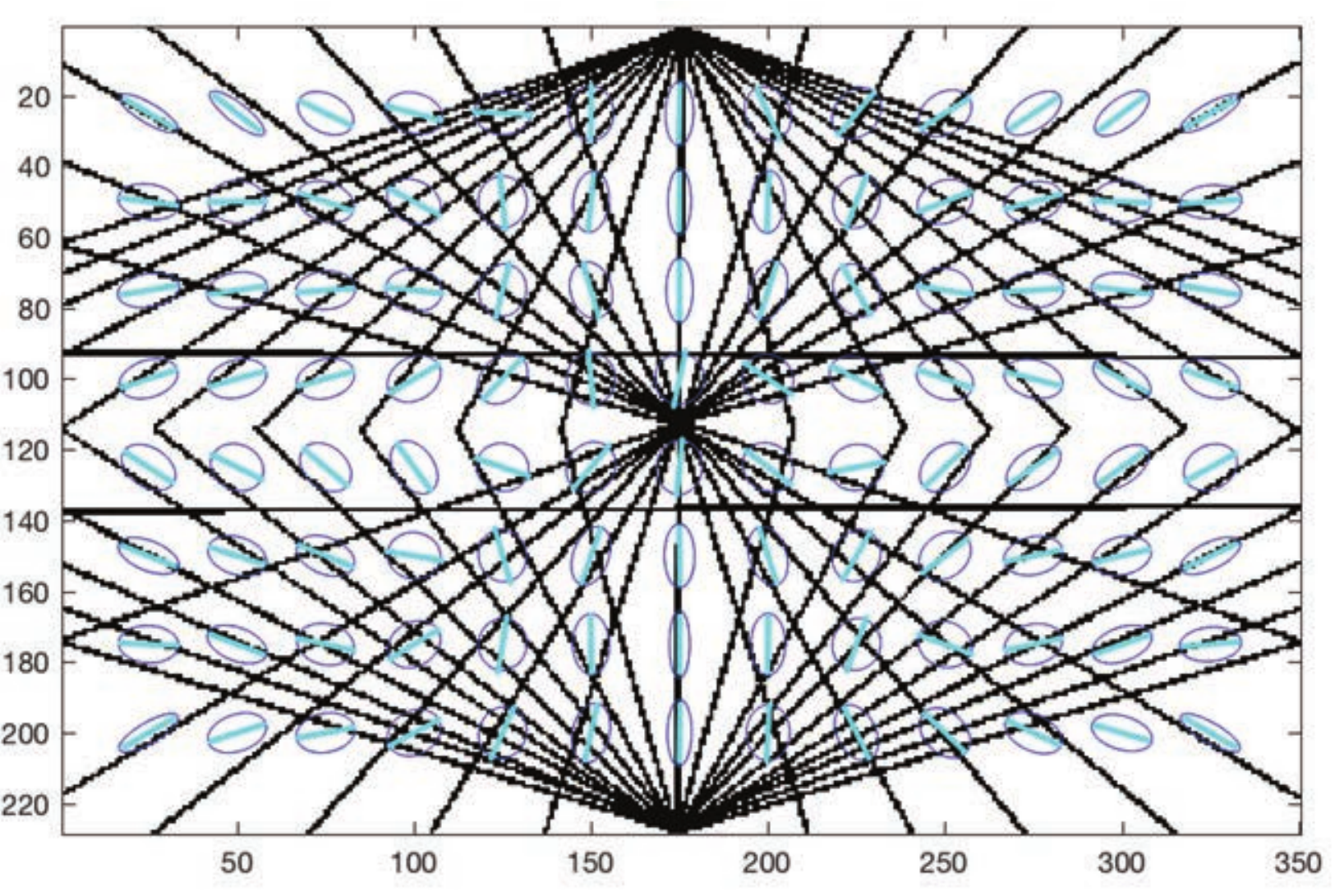}
\caption{}
\label{fig:24:2}
\end{subfigure}
\begin{subfigure}[b]{1.6 in}
\includegraphics[width=\columnwidth]{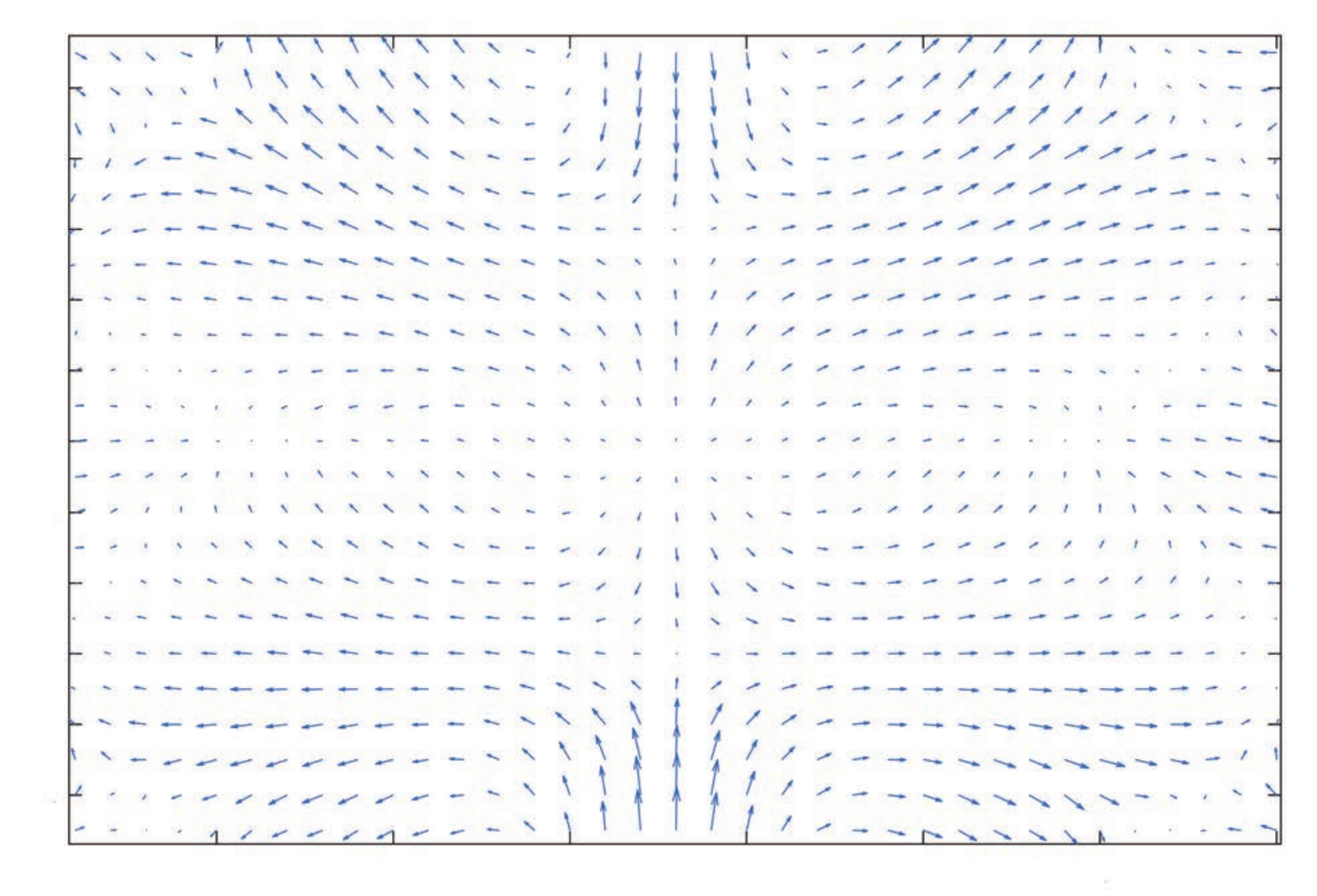}
\caption{}
\label{fig:24:3}
\end{subfigure}
\begin{subfigure}[b]{1.6 in}
\includegraphics[width=\columnwidth]{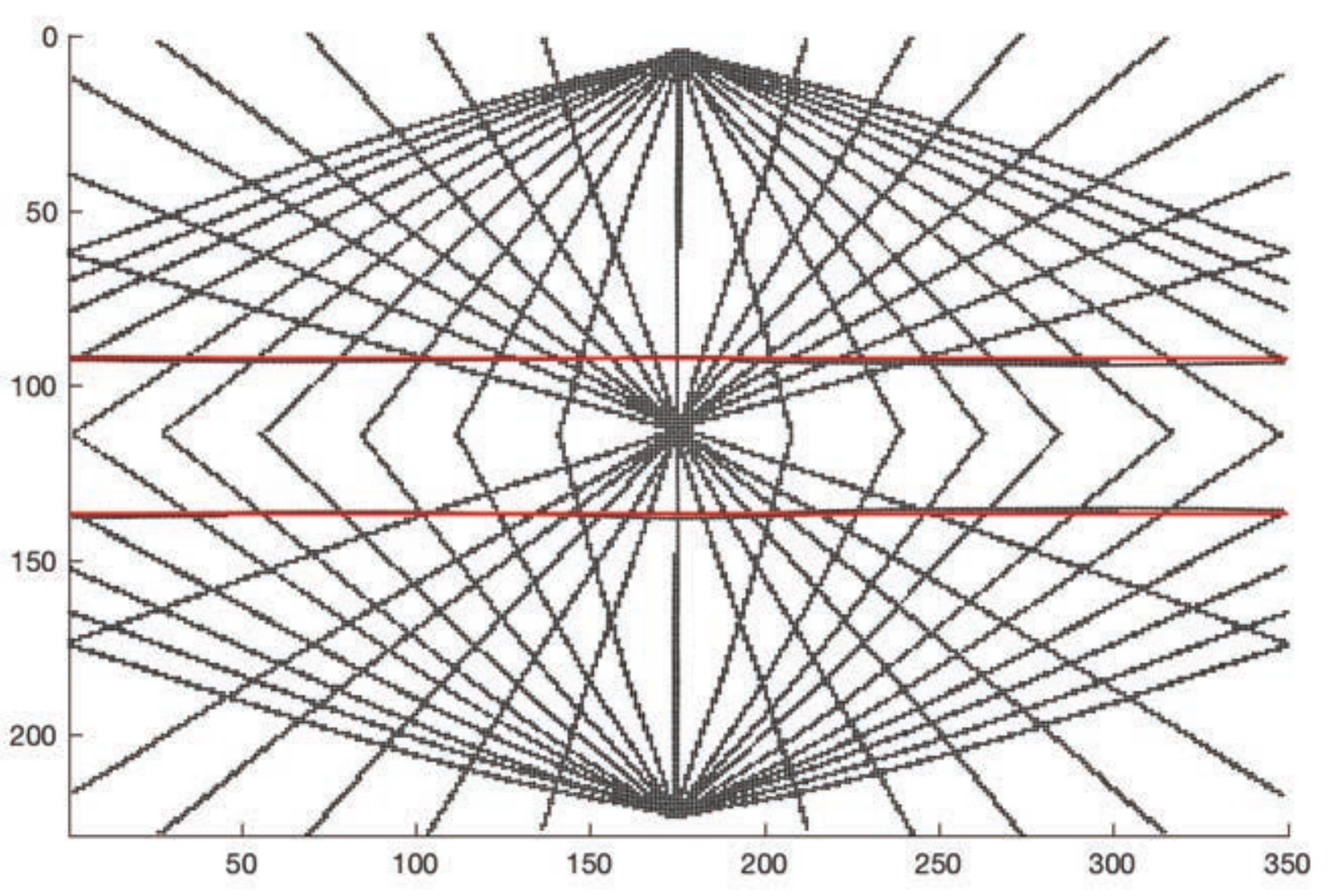}
\caption{}
\label{fig:24:4}
\end{subfigure}
\caption{\subref{fig:24:1} Here we superimpose two red horizontal lines to the original Wundt-Hering illusion shown in \ref{fig:1:6}. \subref{fig:24:2} Representation of $\textbf{p}^{-1}$, projection onto the retinal plane of the polarized connectivity in \ref{ptensor}. The first eigenvalue has direction tangent to the level lines of the distal stimulus. In blue the tensor field, in cyan the eigenvector related to the first eigenvalue. \subref{fig:24:3} Computed displacement field $\bar{u}$. \subref{fig:24:4} Displacement applied to the image. In black we represent the proximal stimulus as displaced points of the distal stimulus: $(x_1, x_2) + \bar{u} (x_1, x_2)$. In red we give two straight lines as reference, in order to put in evidence the curvature of the target lines}
\end{figure} 

\subsection{Z\"{o}llner illusion} \label{sec:36}
\begin{figure}[H]
\centering
\begin{subfigure}[b]{1.0 in}
\centering
\includegraphics[width=\columnwidth]{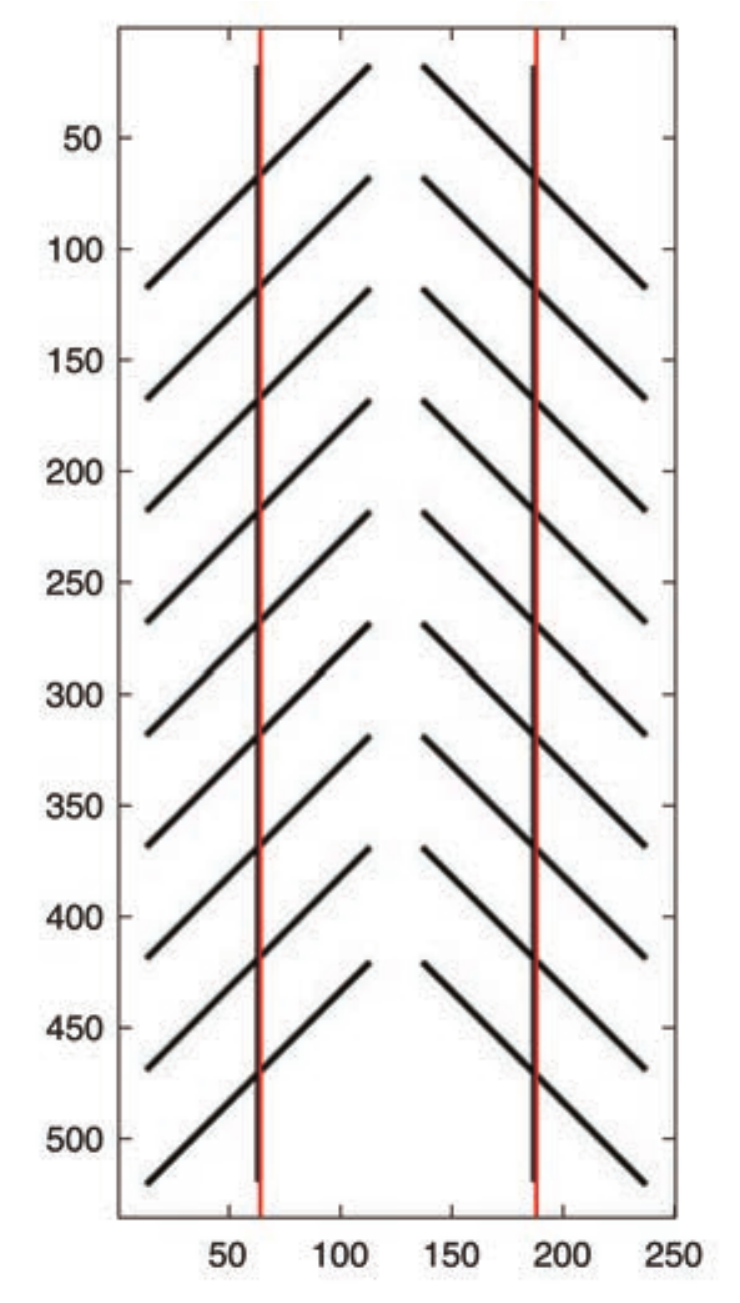}
\caption{}
\label{fig:27:1}
\end{subfigure}
\begin{subfigure}[b]{1.0 in}
\centering
\includegraphics[width=\columnwidth]{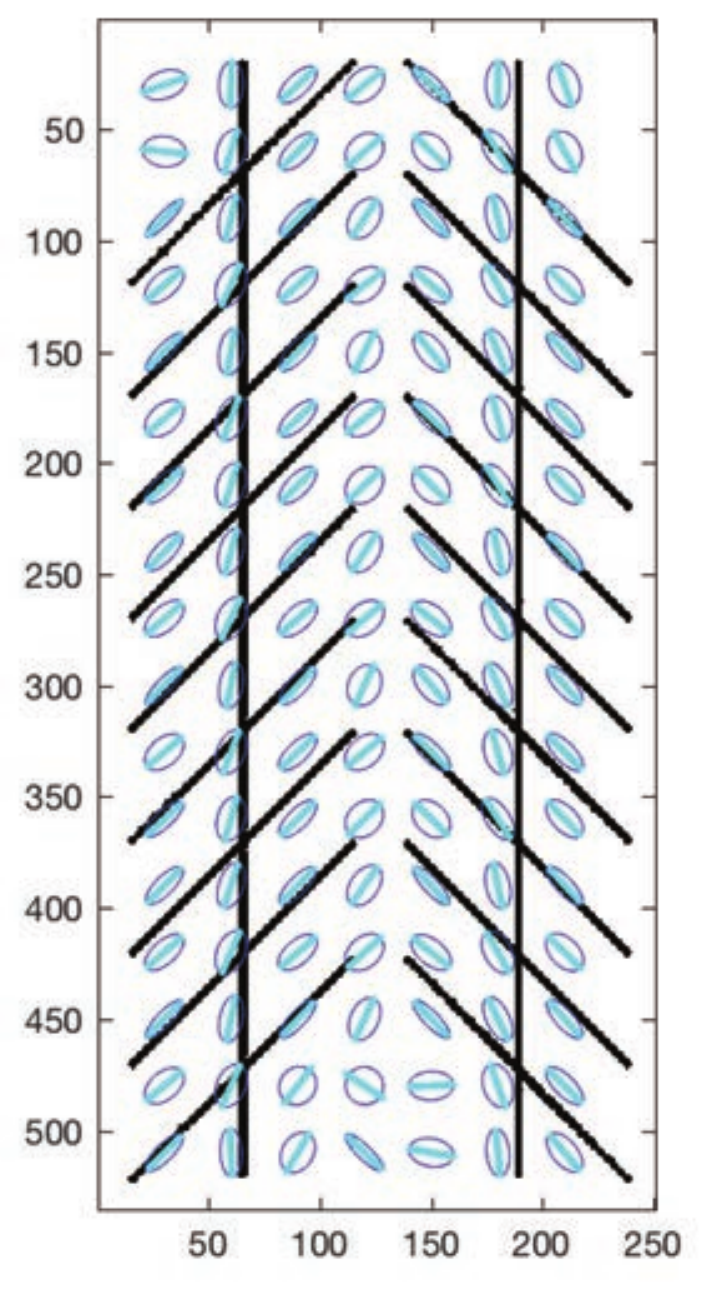}
\caption{}
\label{fig:27:2}
\end{subfigure} \\
\begin{subfigure}[b]{1.0 in}
\centering
\includegraphics[width=\columnwidth]{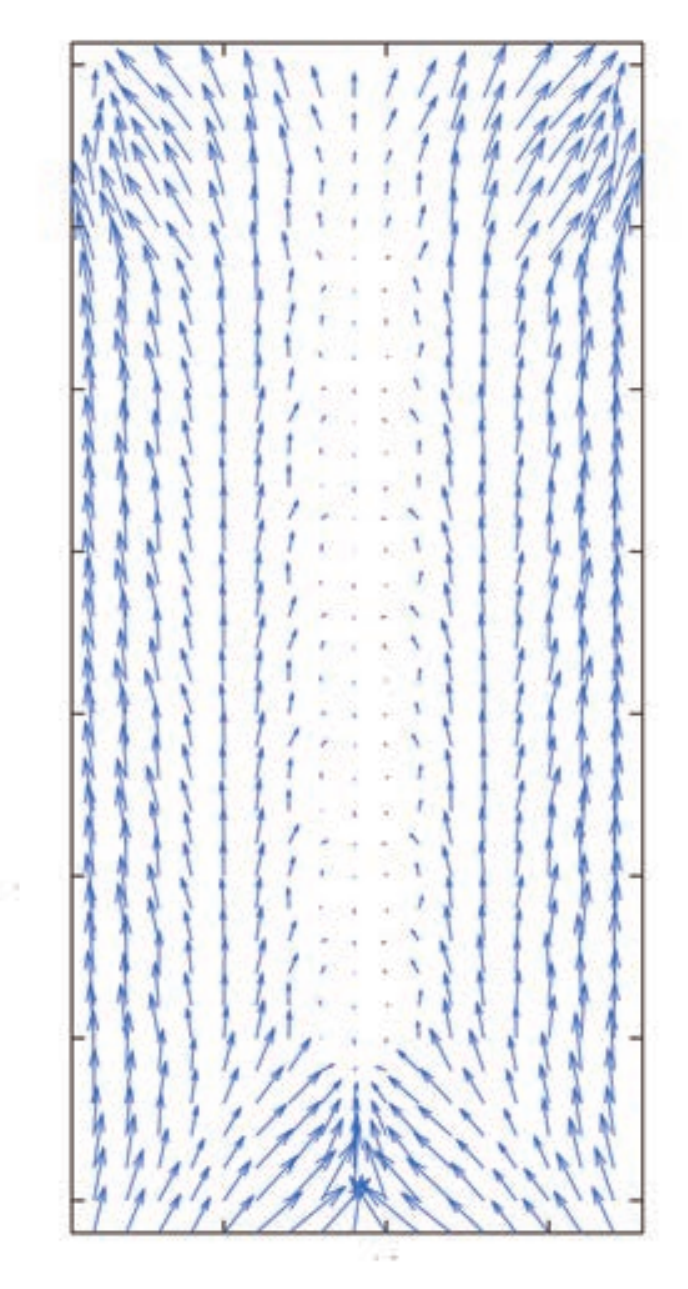}
\caption{}
\label{fig:27:3}
\end{subfigure}
\begin{subfigure}[b]{1.0 in}
\centering
\includegraphics[width=0.93\columnwidth]{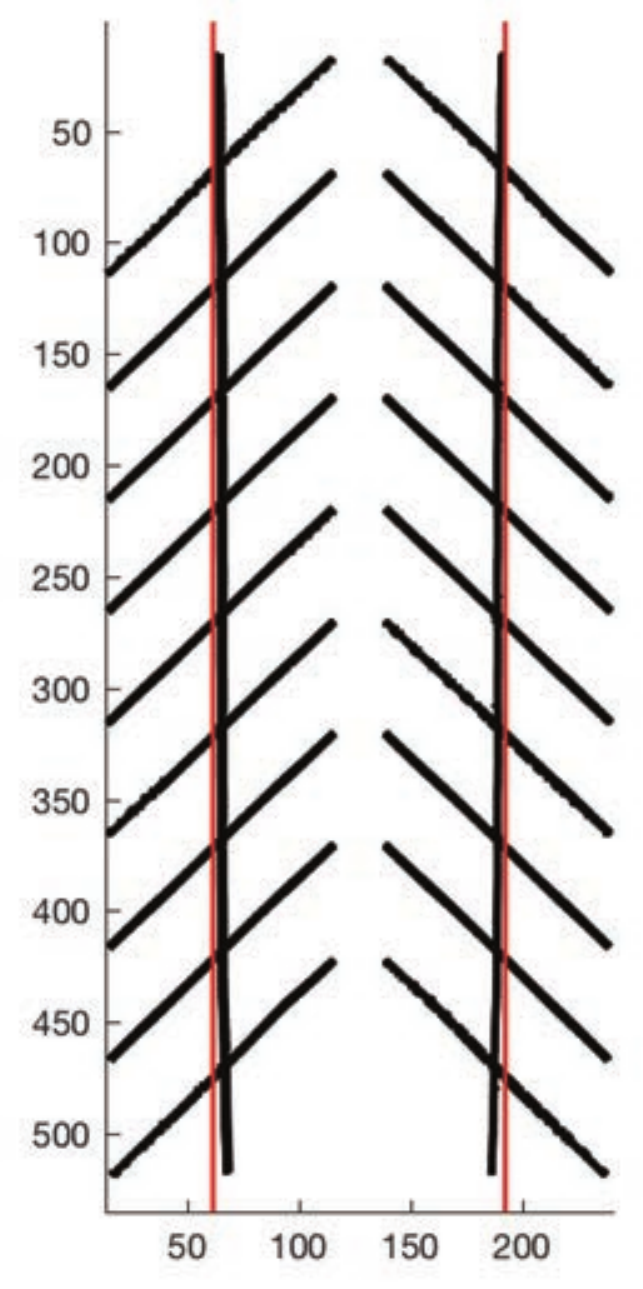}
\label{fig:27:4}
\end{subfigure}
\caption{\subref{fig:27:1} Here we superimpose two red horizontal lines to the original Zollner Illusion shown in \ref{fig:1:2}. \subref{fig:27:2} Representation of $\textbf{p}^{-1}$, projection onto the retinal plane of the polarized connectivity in \ref{ptensor}. The first eigenvalue has direction tangent to the level lines of the distal stimulus. In blue the tensor field, in cyan the eigenvector related to the first eigenvalue. \subref{fig:27:3} Computed displacement field $\bar{u}$. \subref{fig:27:4} Displacement applied to the image. In black we represent the proximal stimulus as displaced points of the distal stimulus: $(x_1, x_2) + \bar{u} (x_1, x_2)$. In red we give two straight lines as reference, in order to put in evidence the unparallelism of the target lines}
\end{figure}

The Z\"{o}llner illusion (figure \ref{fig:1:2}) consists in a pattern of oblique segments surrounding parallel lines, which creates the effect of unparallelism. As in the previous experiments, in figure \ref{fig:27:1} we superimpose two red lines to identify the straight lines. Here we take 32 orientations selected in the interval $[0,\pi)$, $\sigma = 10.08$ pixels. 

\section{Conclusions}
\label{sec:4}
In this paper we presented a neuro-mathematical model based on the functional architecture of the visual cortex to explain and simulate perceptual distortion due to geometrical-optical illusions and to embed geometrical context. 
In our model perceptual distortion is due  to the Riemannian metric induced on the image plane by the connectivity activated by the image stimulus. Its inverse is interpreted as a strain tensor and we computed the deformation in terms of displacement field which arises as solution of \eqref{laplac}.
This technique has been applied to a number of test cases and results are qualitatively in good agreement with human perception. 
In the future this work could be extended to functional architectures involving the feature of scale, starting from models provided by Sarti, Citti and Petitot in \cite{sarti2008symplectic}, \cite{sarti2009functional}. This will allow to provide a model for scale illusions, such as the Delbouf, see \cite{colman2015dictionary}.
Indeed, another direction for future works will be to provide a quantitative analysis for the described phenomena, such as the one proposed by Smith \cite{smith1978descriptive} and to direct compare the developed theory with observations of GOIs through neuro-imaging techniques.

\begin{acknowledgements}
This project has received funding from
the European Union’s Seventh Framework Programme, Marie
Curie Actions- Initial Training Network, under grant agreement
n. 607643, ``Metric Analysis For Emergent Technologies
(MAnET)''. We would like to thank B. ter Haar Romeny, University of Technology Eindhoven, and M. Ursino, University of Bologna, for their important comments and remarks to the present work.
\end{acknowledgements}


\begin{thebibliography}{10}
\providecommand{\url}[1]{{#1}}
\providecommand{\urlprefix}{URL }
\expandafter\ifx\csname urlstyle\endcsname\relax
  \providecommand{\doi}[1]{DOI~\discretionary{}{}{}#1}\else
  \providecommand{\doi}{DOI~\discretionary{}{}{}\begingroup
  \urlstyle{rm}\Url}\fi

\bibitem{angelucci2002circuits}
Angelucci, A., Levitt, J.B., Walton, E.J., Hupe, J.M., Bullier, J., Lund, J.S.:
  Circuits for local and global signal integration in primary visual cortex.
\newblock The Journal of Neuroscience \textbf{22}(19), 8633--8646 (2002)

\bibitem{bosking1997orientation}
Bosking, W.H., Zhang, Y., Schofield, B., Fitzpatrick, D.: Orientation
  selectivity and the arrangement of horizontal connections in tree shrew
  striate cortex.
\newblock The Journal of neuroscience \textbf{17}(6), 2112--2127 (1997)

\bibitem{CS1}
Citti, G., Sarti, A.: A cortical based model of perceptual completion in the
  roto-translation space.
\newblock Journal of Mathematical Imaging and Vision \textbf{24}(3), 307--326
  (2006)

\bibitem{colman2015dictionary}
Colman, A.M.: A dictionary of psychology.
\newblock Oxford University Press, USA (2015)

\bibitem{coren1978seeing}
Coren, S., Girgus, J.S.: Seeing is deceiving: The psychology of visual
  illusions.
\newblock Lawrence Erlbaum (1978)

\bibitem{Daug}
Daugman, J.G.: Uncertainty relation for resolution in space, spatial frequency,
  and orientation optimized by two-dimensional visual cortical filters.
\newblock JOSA A \textbf{2}(7), 1160--1169 (1985)

\bibitem{deangelis1995receptive}
DeAngelis, G.C., Ohzawa, I., Freeman, R.D.: Receptive-field dynamics in the
  central visual pathways.
\newblock Trends in neurosciences \textbf{18}(10), 451--458 (1995)

\bibitem{eagleman2001visual}
Eagleman, D.M.: Visual illusions and neurobiology.
\newblock Nature Reviews Neuroscience \textbf{2}(12), 920--926 (2001)

\bibitem{ehm2012modeling}
Ehm, W., Wackermann, J.: Modeling geometric--optical illusions: A variational
  approach.
\newblock Journal of Mathematical Psychology \textbf{56}(6), 404--416 (2012)

\bibitem{favali2015local}
Favali, M., Citti, G., Sarti, A.: Local and global gestalt laws: A neurally
  based spectral approach.
\newblock Neural computation  (In publication 2016)

\bibitem{Ferm}
Ferm{\"u}ller, C., Malm, H.: Uncertainty in visual processes predicts
  geometrical optical illusions.
\newblock Vision research \textbf{44}(7), 727--749 (2004)

\bibitem{geisler2002illusions}
Geisler, W.S., Kersten, D.: Illusions, perception and bayes.
\newblock Nature neuroscience \textbf{5}(6), 508--510 (2002)

\bibitem{gibson1960concept}
Gibson, J.J.: The concept of the stimulus in psychology.
\newblock American Psychologist \textbf{15}(11), 694 (1960)

\bibitem{von2005treatise}
von Helmholtz, H., Southall, J.P.C.: Treatise on physiological optics, vol.~3.
\newblock Courier Corporation (2005)

\bibitem{Her_1}
Hering, H.E.: Beitr{\"a}ge zur physiologie.
\newblock 1-5 (1861)

\bibitem{hoffman1971visual}
Hoffman, W.C.: Visual illusions of angle as an application of lie
  transformation groups.
\newblock Siam Review \textbf{13}(2), 169--184 (1971)

\bibitem{hubel1977ferrier}
Hubel, D.H., Wiesel, T.N.: Ferrier lecture: Functional architecture of macaque
  monkey visual cortex.
\newblock Proceedings of the Royal Society of London B: Biological Sciences
  \textbf{198}(1130), 1--59 (1977)

\bibitem{jones1987evaluation}
Jones, J.P., Palmer, L.A.: An evaluation of the two-dimensional gabor filter
  model of simple receptive fields in cat striate cortex.
\newblock Journal of neurophysiology \textbf{58}(6), 1233--1258 (1987)

\bibitem{jost2008riemannian}
Jost, J.: Riemannian geometry and geometric analysis.
\newblock Springer Science \&amp; Business Media (2008)

\bibitem{kennedy1985receptive}
Kennedy, H., Martin, K., Orban, G., Whitteridge, D.: Receptive field properties
  of neurones in visual area 1 and visual area 2 in the baboon.
\newblock Neuroscience \textbf{14}(2), 405--415 (1985)

\bibitem{knill1996perception}
Knill, D.C., Richards, W.: Perception as Bayesian inference.
\newblock Cambridge University Press (1996)

\bibitem{koenderink1990receptive}
Koenderink, J.J., Van~Doorn, A.: Receptive field families.
\newblock Biological cybernetics \textbf{63}(4), 291--297 (1990)

\bibitem{koffka2013principles}
Koffka, K.: Principles of Gestalt psychology, vol.~44.
\newblock Routledge (2013)

\bibitem{levitt1994receptive}
Levitt, J.B., Kiper, D.C., Movshon, J.A.: Receptive fields and functional
  architecture of macaque v2.
\newblock Journal of Neurophysiology \textbf{71}(6), 2517--2542 (1994)

\bibitem{lubliner2008plasticity}
Lubliner, J.: Plasticity theory.
\newblock Courier Corporation (2008)

\bibitem{marsden1994mathematical}
Marsden, J.E., Hughes, T.J.: Mathematical foundations of elasticity.
\newblock Courier Corporation (1994)

\bibitem{murray2013illusory}
Murray, M.M., Herrmann, C.S.: Illusory contours: a window onto the
  neurophysiology of constructing perception.
\newblock Trends in cognitive sciences \textbf{17}(9), 471--481 (2013)

\bibitem{murray2002spatiotemporal}
Murray, M.M., Wylie, G.R., Higgins, B.A., Javitt, D.C., Schroeder, C.E., Foxe,
  J.J.: The spatiotemporal dynamics of illusory contour processing: combined
  high-density electrical mapping, source analysis, and functional magnetic
  resonance imaging.
\newblock The Journal of Neuroscience \textbf{22}(12), 5055--5073 (2002)

\bibitem{oppel1855uber}
Oppel, J.J.: Uber geometrisch-optische tauschungen.
\newblock Jahresbericht des physikalischen Vereins zu Frankfurt am Main  (1855)

\bibitem{petitot2008neurogeometrie}
Petitot, J.: Neurog{\'e}om{\'e}trie de la vision (2008)

\bibitem{robinson2013psychology}
Robinson, J.O.: The psychology of visual illusion.
\newblock Courier Corporation (2013)

\bibitem{sanguinetti2010model}
Sanguinetti, G., Citti, G., Sarti, A.: A model of natural image edge
  co-occurrence in the rototranslation group.
\newblock Journal of vision \textbf{10}(14), 37--37 (2010)

\bibitem{sarti2008symplectic}
Sarti, A., Citti, G., Petitot, J.: The symplectic structure of the primary
  visual cortex.
\newblock Biological Cybernetics \textbf{98}(1), 33--48 (2008)

\bibitem{sarti2009functional}
Sarti, A., Citti, G., Petitot, J.: Functional geometry of the horizontal
  connectivity in the primary visual cortex.
\newblock Journal of Physiology-Paris \textbf{103}(1), 37--45 (2009)

\bibitem{smith1978descriptive}
Smith, D.A.: A descriptive model for perception of optical illusions.
\newblock Journal of Mathematical Psychology \textbf{17}(1), 64--85 (1978)

\bibitem{song2013effective}
Song, C., Schwarzkopf, D.S., Lutti, A., Li, B., Kanai, R., Rees, G.: Effective
  connectivity within human primary visual cortex predicts interindividual
  diversity in illusory perception.
\newblock The Journal of Neuroscience \textbf{33}(48), 18,781--18,791 (2013)

\bibitem{von1984illusory}
Von Der~Heyclt, R., Peterhans, E., Baurngartner, G.: Illusory contours and
  cortical neuron responses.
\newblock Science \textbf{224} (1984)

\bibitem{wade1982art}
Wade, N.: The art and science of visual illusions.
\newblock Routledge Kegan \& Paul (1982)

\bibitem{walker1973mathematical}
Walker, E.H.: A mathematical theory of optical illusions and figural
  aftereffects.
\newblock Perception \&amp; Psychophysics \textbf{13}(3), 467--486 (1973)

\bibitem{weiss2002motion}
Weiss, Y., Simoncelli, E.P., Adelson, E.H.: Motion illusions as optimal
  percepts.
\newblock Nature neuroscience \textbf{5}(6), 598--604 (2002)

\bibitem{westheimer2008illusions}
Westheimer, G.: Illusions in the spatial sense of the eye: Geometrical--optical
  illusions and the neural representation of space.
\newblock Vision research \textbf{48}(20), 2128--2142 (2008)

\bibitem{wundt1898geometrisch}
Wundt, W.M.: Die geometrisch-optischen T{\"a}uschungen, vol.~24.
\newblock BG Teubner (1898)

\bibitem{young1987gaussian}
Young, R.A.: The gaussian derivative model for spatial vision: I. retinal
  mechanisms.
\newblock Spatial vision \textbf{2}(4), 273--293 (1987)

\end{thebibliography}

\end{document}